\documentclass[lettersize,journal,10pt]{IEEEtran}
\usepackage{amsmath,amsfonts}
\usepackage{algorithmic}
\usepackage{algorithm}
\usepackage{array}
\usepackage[caption=false,font=footnotesize,labelfont=rm,textfont=rm]{subfig} 

\usepackage{textcomp}
\usepackage{stfloats}
\usepackage{url}
\usepackage{verbatim}
\usepackage{graphicx}
\usepackage{cite}

\usepackage{xcolor}
\definecolor{StrongPurple}{RGB}{148, 0, 211}

\usepackage{amssymb, bm, mathrsfs,mathtools}
\usepackage{upgreek}
\usepackage{booktabs}
\usepackage{hyperref}
\usepackage{float} 
\usepackage{placeins} 
\usepackage{tabularx}
\usepackage{bbm}
\usepackage{threeparttable} 
\usepackage{multirow}

\usepackage{wasysym}
\newcommand{\specialcell}[2][c]{%
\begin{tabular}[#1]{@{}c@{}}#2\end{tabular}}

\usepackage{tikz}
\newcommand*{\mycirc}[1]{%
  \tikz[baseline=(char.base)]{
    \node[draw,circle,inner sep=1pt] (char) {#1};
  }%
}

\begin{document}

\title{MoTDiff: High-resolution Motion Trajectory estimation from a single blurred image using Diffusion models}

\author{Wontae Choi,
        Jaelin Lee, \IEEEmembership{Student Member, IEEE},
        Hyung Sup Yun,\\
        Byeungwoo Jeon, \IEEEmembership{Senior Member, IEEE}, and Il Yong Chun, \IEEEmembership{Member, IEEE}
        

\thanks{Wontae Choi is with the Department of Artificial Intelligence (AI), Sungkyunkwan University (SKKU), Suwon 16419, South Korea. (email: \href{mailto:wontae1998@g.skku.edu}{wontae1998@g.skku.edu})}

\thanks{Jaelin Lee is with the Department of Electrical and Computer Engineering (ECE), SKKU, Suwon 16419, South Korea. (email: \href{mailto:jaelin@skku.edu}{jaelin@skku.edu})}

\thanks{Hyung Sup Yun is with the New Technology Team, ALLforLAND Co., Ltd., Seoul 07792, South Korea. 
He was with the Department of ECE, SKKU,  Suwon 16419, South Korea.
(email: \href{emailto:hyungsup0225@g.skku.edu}{hyungsup0225@g.skku.edu})}

\thanks{Byeungwoo Jeon is with the Department of ECE, SKKU, Suwon 16419, South Korea. (email: \href{emailto:bjeon@skku.edu}{bjeon@skku.edu})}

\thanks{
Il Yong Chun is with the Departments of 
AI, ECE, 
Semiconductor Convergence Engineering, 
Advanced Display Engineering, 
and Display Convergence Engineering, 
SKKU, Suwon, 16419, South Korea. 
He is also with the Center for Neuroscience Imaging Research, 
Institute for Basic Science, Suwon, 16419, South Korea 
(e-mail: \href{emailto:iychun@skku.edu}{iychun@skku.edu}).
}

\thanks{
\textit{(Wontae Choi and Jaelin Lee are contributed equally to this work. 
Corresponding authors: Hyung Sup Yun; Byeung Woo
Jeon; Il Yong Chun.)}
}
}



\maketitle

\begin{abstract}
Accurate estimation of motion information is crucial in 
diverse computational imaging and computer vision applications.
Researchers have investigated various methods to extract motion information from a single blurred image, including blur kernels and optical flow.
However, existing motion representations are often of low quality, i.e., coarse-grained and inaccurate. 
In this paper, 
we propose the first high-resolution (HR) Motion Trajectory estimation framework using Diffusion models (\emph{MoTDiff}). 
Different from existing motion representations,
we aim to estimate an HR motion trajectory with high-quality from a single motion-blurred image.
The proposed \emph{MoTDiff} consists of two key components: 
\textit{1)} a new conditional diffusion framework that uses multi-scale feature maps extracted from a single blurred image as a condition, and 
\textit{2)} a new training method that can promote precise identification of a fine-grained motion trajectory, consistent estimation of overall shape and position of a motion path, and pixel connectivity along a motion trajectory.
Our experiments demonstrate that the proposed \emph{MoTDiff} can outperform state-of-the-art methods in both blind image deblurring and coded exposure photography applications.
\end{abstract}


\section{Introduction}
\IEEEPARstart{M}{otion} blur occurs when relative movement between a camera and an element/elements of scene causes point sources to spread across the image sensor during exposure. 
Motion blur or motion can be either spatially-variant or invariant.
The spatially-variant blur varies locally across image, 
where its general cause includes object motion(s) and depth variations in the scene (when a camera is fixed).
The spatially-invariant blur is uniform across entire image, and can be in general caused by camera shake or movement.
Spatially-invariant motion has been typically captured by a point spread function (PSF) \cite{ndeblur,selfdeblur,blind-dps,kernel-diff}.

Estimating motion information is crucial in various computational imaging applications including blind image deblurring, coded exposure photography (CEP) \cite{firstCEP,dnfcep}, and event camera.
Estimating motion information can help restore clear images that can improve many computer vision technologies, such as image classification, semantic segmentation, and object detection.
In deblurring spatially-variant motion(s) in a single blurred image, 
researchers have proposed different motion representations. 
Spatially-variant motion representations include classical ones, e.g., PSFs for different pixel locations \cite{densitymotion,depthmotion}, and 
recent ones, e.g., optical flow \cite{OFEst}, pixel-wise parametric trajectory \cite{ETR}, patch/pixel-wise parametric motion vector \cite{FlowSun,FlowGong}, and motion path of an object \cite{sifmo}.
However, the optical flow-based method \cite{OFEst} estimates linear motions between the first and last frames in a single blurred image, 
so its estimated optical flows can inherently only represent linear motion at each pixel.
Similarly, the parametric representations \cite{ETR, FlowSun,FlowGong} have limitations in capturing complex motion patterns.
For example, the parametric motion vector-based methods \cite{FlowSun, FlowGong} model motion by a two-dimensional (2D) motion vector, that is, a straight line, at the patch level \cite{FlowSun} or the pixel level \cite{FlowGong}.
The parametric trajectory-based method \cite{ETR} models per-pixel motion with a continuous trajectory, parameterized using either a linear or a quadratic function.

To deblur spatially-invariant motion in a single blurred image, many optimization methods have been proposed in a blind manner, i.e., by simultaneously estimating a PSF and recovering a latent sharp image, with different kernel assumptions or image prior models \cite{spatialprior,mapreview,gradientprior,pmp}.
However, the estimated PSF often appears noisy.
Recently, researchers have proposed different deep learning approaches for blind and spatially-invariant deblurring \cite{ndeblur,selfdeblur,blind-dps,kernel-diff}.
SelfDeblur uses two generative networks to capture the deep priors of the latent sharp image and the blur kernel in a zero-shot manner \cite{selfdeblur}.
Blind-DPS uses parallel diffusion models to jointly estimate the blur kernel and the latent sharp image \cite{blind-dps}.
Kernel-Diff uses a single diffusion model for estimating a blur kernel, conditioned on an observed blurred image, where an estimated kernel is subsequently used in a non-blind deblurring solver to obtain a clear image from a blurred input \cite{kernel-diff}.
Yet, the motion trajectory captured in an estimated PSF is often of low quality, i.e., it is blurry and/or disconnected (where its ground truth is an uninterrupted motion trajectory).

A high-resolution (HR) motion trajectory may improve performances in computational imaging applications such as image deblurring and CEP \cite{firstCEP,dnfcep}.
For example, HR motion trajectories may improve the motion deblurring performances -- particularly for blurred images with complex motion trajectories -- by capturing fine-grained motion details.
It was shown that the more accurately the PSFs are estimated, the more effectively the latent images can be restored \cite{mnc}.
An HR motion trajectory with fine-grained motion details can facilitate more precise PSF modeling, ultimately enhancing motion deblurring performance.
Another example is CEP that can enhance the invertibility of an imaging system by modulating motion with a code, i.e., shutter fluttering pattern.
Since a code is iteratively optimized along a given motion trajectory \cite{firstCEP,dnfcep}, accurate estimation of the motion trajectory is important for effective code optimization.
Conversely, blurry or low-resolution motion estimates inevitably compromise its optimality.
For effective code optimization in CEP, to accurately estimate a high-resolution motion trajectory is critical.
\footnote{As an alternative, one can use hardware sensors such as gyroscopes embedded in cameras, accelerometers, or laser-reflector systems to directly obtain motion information \cite{firstCEP}.
Yet, motion information obtained from these sensors may not accurately capture complex 2D motion trajectories on the image plane. 
This limitation arises from measurement noise and the difficulty of converting raw sensor data into precise pixel-level motion \cite{gyroerror}, ultimately resulting in sub-optimal code generation \cite{dnfcep}.}

This paper proposes the \emph{first} conditional diffusion model that can estimate an accurate HR motion trajectory directly from a single motion-blurred image, referred to as the {\bfseries Motion Trajectory Diffusion model (MoTDiff).}
The proposed framework has the following contributions:
\begin{itemize}

    \item {\bfseries New conditional diffusion model, MoTDiff:} 
    We propose a new conditioning approach for diffusion models.
    In particular, we extract multi-scale motion features from a blurred image using the Pyramid Vision Transformer (PVT) architecture \cite{pvt},
    where we observed that a deep PVT stage can capture the semantic context of a motion trajectory embedded in a blurred image.
    We then adapt a stepwise adaptive method to aggregate high-level motion representations from deep PVT stages to low-level motion features from early PVT stages. 
    We use aggregated features as guidance/a condition in a diffusion model.

    \item {\bfseries New training method for MoTDiff:}    
    We propose a new training loss function that can encourage precise identification of a fine-grained motion path and consistently estimate the overall shape and position of the target HR motion trajectory.        
    In addition, we propose a new training strategy that can promote the connectivity of a motion trajectory.
    
    \item {\bfseries Superior performances in two computational imaging applications:} Our experiments with two computational imaging applications, blind image deblurring and CEP, demonstrate that the proposed framework outperforms state-of-the-art (SOTA) methods in each application.

\end{itemize}

\section{Backgrounds}

\subsection{Conditional diffusion models}
\label{sec:cond-dm}

Denoising Diffusion Probabilistic Models (DDPM) \cite{ddpm, sr3,deblur_refine,repaint,palette,medsegdiff,ddlfrm,dx2ct,precipitation, lotus, v-pred} are a class of generative methods, characterized by Markov chains of forward and reverse diffusion processes.
The forward process that iteratively adds isotropic Gaussian noise to an original sample \( \mathbf{x}_0 \) is defined as follows:
\begin{equation}
\mathbf{x}_t = \sqrt{\alpha_t}\mathbf{x}_0 + \sqrt{1 - \alpha_t}\boldsymbol{\upepsilon},
\label{eq:forward_process}
\end{equation}
where \( \mathbf{x}_t \) and \( \alpha_t \) represent corrupted sample and a constant determined by a noise schedule at the timestep \( t = 1, \ldots, T \), respectively, and \( \boldsymbol{\upepsilon} \sim \mathcal{N}(\mathbf{0}, \mathbf{I}) \). The reverse process is a denoising process that denoises a noisy sample \( \mathbf{x}_t \) to \( \mathbf{x}_{t-1} \). 
Starting from \( \mathbf{x}_T \sim \mathcal{N}(\mathbf{0}, \mathbf{I}) \), this process is iteratively performed until the clean sample \( \mathbf{x}_0 \) is generated using the trained denoiser \( D_{\boldsymbol{\uptheta}}( \mathbf{x}_t, t ) \). 
Specifically, \( D_{\boldsymbol{\uptheta}}( \mathbf{x}_t, t ) \) is a model with parameters \({\uptheta}\) that, at each timestep \(t\), predicts one of the followings from $\mathbf{x}_t$: (i) the added noise \(\boldsymbol{\upepsilon}_{t}\) \cite{ddpm,sr3,deblur_refine, repaint, palette, medsegdiff, ddlfrm, dx2ct}, (ii) the original sample \( \mathbf{x}_0 \) \cite{precipitation,lotus}, or (iii) the linear combination of $\boldsymbol{\upepsilon}_{t}$ and $\mathbf{x}_0$, $\mathbf{v}_t = \sqrt{\alpha_t} \boldsymbol{\upepsilon}_t + \sqrt{1 - \alpha_t} \, \mathbf{x}_0$ \cite{v-pred}.

In conditional diffusion models \cite{sr3,deblur_refine,repaint,palette,medsegdiff,ddlfrm,dx2ct,precipitation,lotus}, the reverse process can incorporate additional guidance information so-called a \emph{condition.} 
Depending on the given condition(s) and conditioning method, one can manipulate generation results.

\subsection{PVT}

The Vision Transformer (ViT) architecture
\cite{ViT} uses the attention mechanism \cite{attention} for vision tasks and achieved high performances in diverse vision applications.
However, ViT \cite{attention} processes the input image with a single scale that may restrict its ability to capture fine-grained details and global context, particularly useful for dense prediction tasks, such as object detection and segmentation.
To address this, 
PVT \cite{pvt} uses a progressive shrinking strategy where each stage uses a patch embedding layer with different patch sizes to create multi-scale feature maps.
These embeddings are used in spatial-reduction attention that reduces the size of key value embeddings to efficiently process \ feature maps and reduce computation memory costs.
The transformer encoder takes position-embedded patches as an input and produces the multi-scale features via the above attention process.

\section{Methods}

This section introduces the proposed MoTDiff framework that estimates an HR motion trajectory from a motion-blurred image. 
Section~\ref{sec:HR motion trajectory} describes our target, HR motion trajectories.
Section~\ref{sec:overall_architecture} provides an overview of MoTDiff and explains its network architecture.
Section~\ref{sec:loss} explains the proposed training loss function for MoTDiff;
Section~\ref{sec:STPD} introduces the proposed training strategy that promotes the connectivity of pixels in an HR motion trajectory.

\begin{figure}[t] 
\centering
\subfloat[\label{fig:HFtraj_dot}]{\includegraphics[width=0.24\columnwidth]{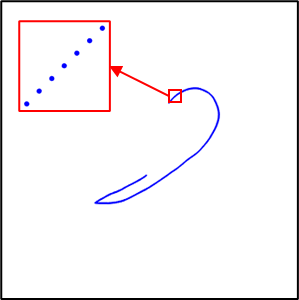}}\hspace{-0.01cm}
\subfloat[\label{fig:HFtraj_LFtraj}]{\includegraphics[width=0.24\columnwidth]{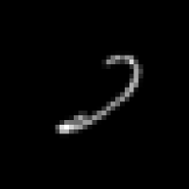}}\hspace{-0.01cm}
\subfloat[\label{fig:HFtraj_vector}]{\includegraphics[width=0.24\columnwidth]{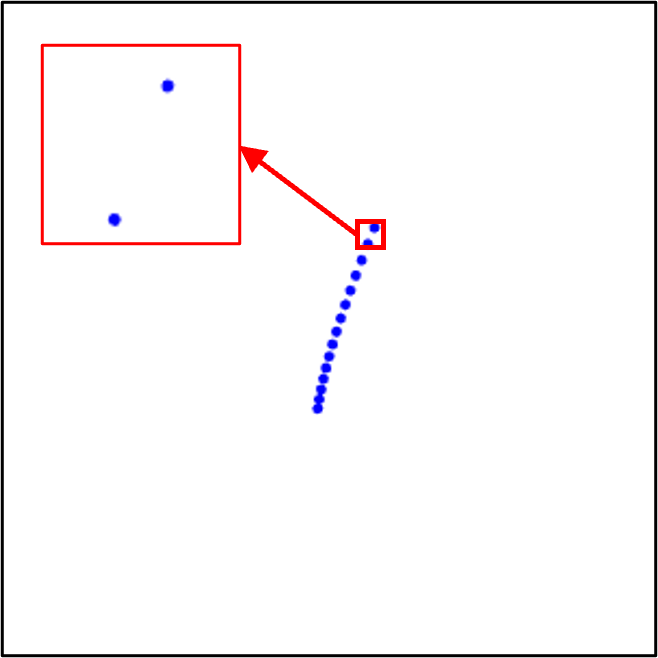}}\hspace{-0.01cm}
\subfloat[\label{fig:HFtraj_HFtraj}]{\includegraphics[width=0.24\columnwidth]{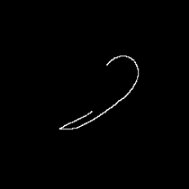}}

\caption{Illustrations of different motion trajectory representations for the same 2D motion path. 
(a) A set of trajectory positions \cite{trajmodeling} (continuous space). 
(b) PSF \cite{blind-dps,kernel-diff} (discrete space with $64\!\times\!64$ pixels).
(c) Parametric trajectory \cite{ETR} (quadratic curve constraint; continuous space). 
(d) Proposed HR trajectory (discrete space with $256\!\times\!256$ pixels).}

\label{fig:HFtraj}
\end{figure}

\subsection{HR motion trajectory representation}
\label{sec:HR motion trajectory}

In motion processing tasks, there exist various motion trajectory representations and models \cite{blind-dps,kernel-diff,ETR,trajmodeling,DeblurGan,SelfSupervised}.
We describe different trajectory representations and their limitations:

\begin{itemize}
\item {\bfseries A set of motion trajectory positions:} 
The trajectory representation introduced in \cite{trajmodeling}, 
later adopted in \cite{DeblurGan, SelfSupervised}, 
is a set of motion trajectory positions defined over a continuous space.
See Fig.~\ref{fig:HFtraj_dot}. 
Estimating coordinates of trajectory points in continuous space poses challenges such as 
sensitivity to noise,
difficulty in designing loss function to capture structural consistency, and
discretization gap.
First, direct regression of continuous coordinates makes models vulnerable to minor perturbations and inherent data biases. 
This can hinder robust and generalizable learning.
Second, it is challenging to design loss functions that can capture high-level structural consistency, 
such as trajectory or shape alignment.
Basic loss functions such as mean squared error and mean absolute error are limited in capturing structural consistency.
Third, final estimates are often required to be mapped back to a discrete pixel grid, 
where the continuous-to-discrete conversion may introduce artifacts such as subpixel misalignments and aliasing.

\item {\bfseries PSF:} 
A PSF is produced by interpolating a set of trajectory points \cite{blind-dps, kernel-diff}.
A PSF is in general defined over a coarser pixel grid compared to the resolution of an input blurred image.
See Fig.~\ref{fig:HFtraj_LFtraj}.
Thus, the PSF lacks the resolution necessary to represent subtle subpixel motion and the fine structure of complex trajectories.
In addition, multiple distinct motion trajectories can produce similar PSFs, leading to an ill-posed inverse problem.

\item {\bfseries Parametric trajectory:}
The parametric trajectory representation is a set of motion trajectory points defined over a continuous space that conforms to a predefined motion pattern, such as linear or quadratic curves \cite{ETR}.
See Fig.~\ref{fig:HFtraj_vector}.
This method is challenging in capturing complex or non-linear motion patterns, as it relies on a simple predefined constraint. 
As a result, it may fail to represent realistic motion trajectories that involve abrupt changes, curved paths, or fine-grained variations.

\item {\bfseries Proposed HR motion trajectory:}
We define an HR trajectory by mapping a set of trajectory points to a pixel grid with the same spatial resolution as the input blurred image.
We scale the coordinates of a set of trajectory positions and map them to the HR pixel grid.
See Fig.~\ref{fig:HFtraj_HFtraj}.
The proposed HR motion trajectory representation can resolve the limitations of the aforementioned existing representations.
We expect that the proposed representation can fully recover the underlying motion characteristics such as direction and curvature by identifying fine-grained or complex motion patterns.
\end{itemize}

The aim of our research is to estimate an HR motion trajectory \emph{directly} from an observed motion-blurred image.

\subsection{Proposed conditional diffusion models, MoTDiff}
\label{sec:overall_architecture}

\begin{figure*}[t!]
\centering
\centerline{\includegraphics[width=0.9\textwidth]{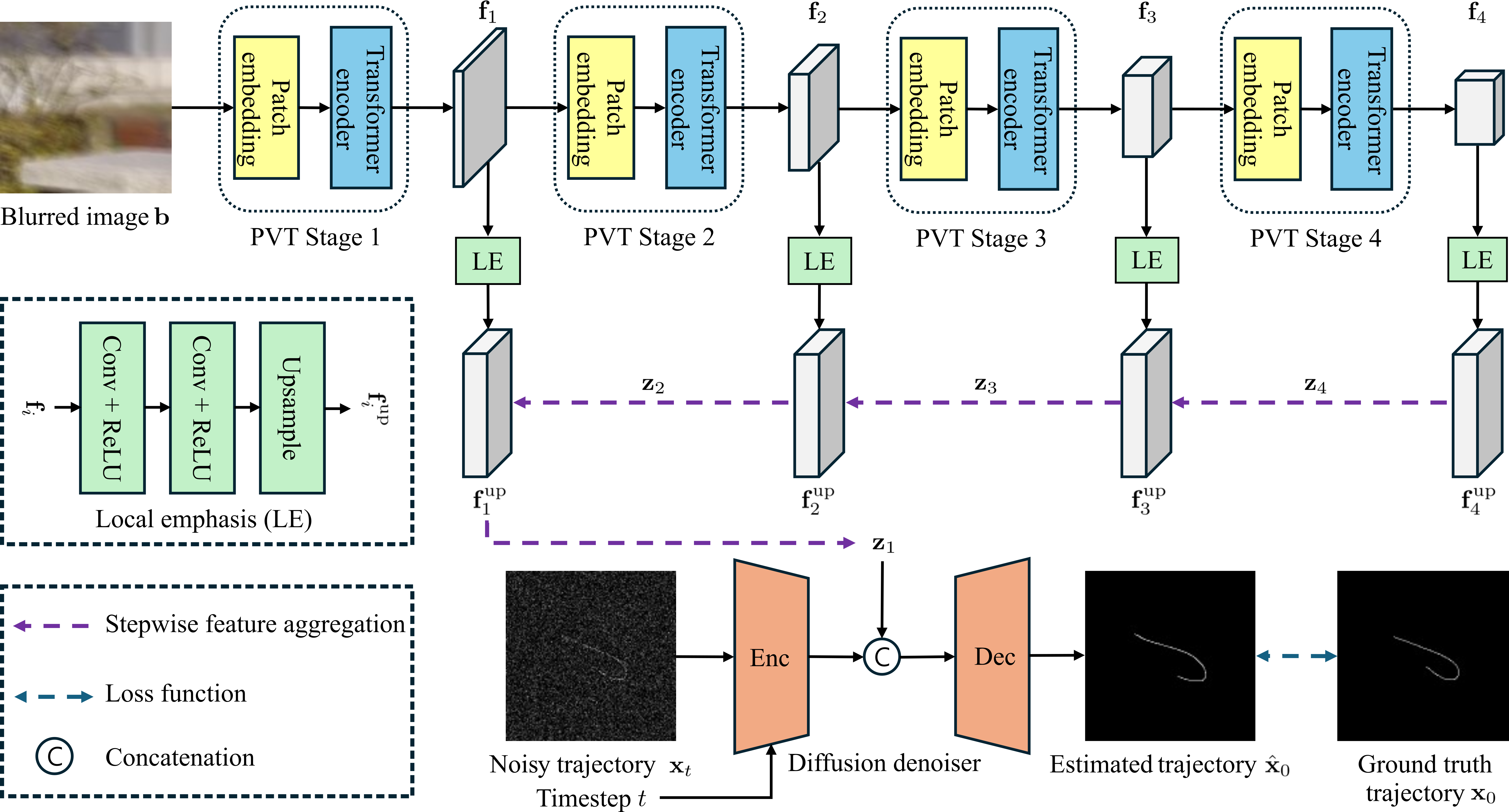}}

\caption{Overview of proposed {\bfseries MoTDiff} in the reverse diffusion process.
To extract a condition in MoTDiff, we first extract multi-scale feature maps $\{ \mathbf{f}_s \}$ from a single blurred image $\mathbf{b}$ using {\bfseries PVT}.
We then enhance salient motion features in $\{ \mathbf{f}_s \}$ ({\bfseries local emphasis}), and progressively integrate global and local motion features \( \{ \mathbf{f}^{\text{up}}_s \} \) across the feature hierarchy ({\bfseries stepwise feature aggregation}).
We use the aggregated feature map $\mathbf{z}_1$ as a condition for a diffusion denoiser that gives a motion trajectory estimate $\hat{\mathbf{x}}_0$ from noisy trajectory $\mathbf{x}_t$ at sampling timestep $t$, $\forall t$.
We train MoTDiff using loss functions that compare an estimated trajectory $\hat{\mathbf{x}}_0$ with the ground truth  $\mathbf{x}_0$, for uniformly randomly sampled timesteps.
}
\label{fig:network}
\end{figure*}

For HR motion trajectory estimation, 
we propose a new conditioning approach on top of the conditional DDPM framework (see Section~\ref{sec:cond-dm}).
Specifically, we generate multi-scale features from a motion-blurred image and sophisticatedly integrate them as a condition into encoded features from a diffusion denoiser.
The proposed conditioning approach can
identify latent motion information in a blurred image and guide diffusion U-Net to generate a fine-grained and accurate trajectory.
Fig.~\ref{fig:network} illustrates the overview of the reverse process of proposed MoTDiff.

\subsubsection{Proposed conditioning approach: Multi-scale feature extraction}
\label{subsubsec:feature extraction}

From a motion-blurred image \( \mathbf{b} \in \mathbb{R}^{H \times W \times 3}\), 
we adapt the PVT architecture \cite{pvt} of which different PVT stages extract feature maps with different scales,
and capture multi-scale features hierarchically:
\begin{equation}
\mathcal{F}_\text{multi-scale} 
= 
\{\mathbf{f}_s : s = 1,\ldots,4  \} 
= 
\text{PVT}_{\boldsymbol{\xi}}(\mathbf{b}),
\label{eq:feat-ms}
\end{equation}
where \( \mathbf{f}_s \in \mathbb{R}^{H_s \times W_s \times C_s} \) denotes the feature map obtained from the \( s \)th PVT stage, and $\boldsymbol{\xi}$ is the parameters of PVT.

\begin{figure}[t] 
\centering
\subfloat[\label{fig:Blurred}]{\includegraphics[width=0.24\columnwidth]{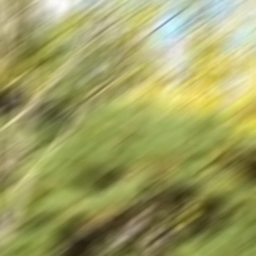}}\hspace{-0.01cm}
\subfloat[\label{fig:Traj}]{\includegraphics[width=0.24\columnwidth]{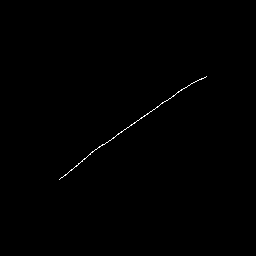}}\hspace{-0.01cm}
\subfloat[\label{fig:FeatureMap1}]{\includegraphics[width=0.24\columnwidth]{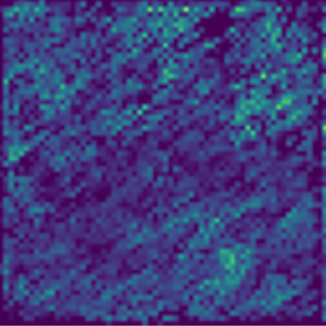}}\hspace{-0.01cm}
\subfloat[\label{fig:FeatureMap4}]{\includegraphics[width=0.24\columnwidth]{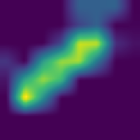}}

\caption{
Visualizations of motion features with different levels of understanding captured via PVT (PVT Stages 1 \& 4).
(a) An input motion-blurred image to the PVT encoder ($256\times256$ pixels). 
(b) Ground-truth HR motion trajectory ($256\times256$ pixels). 
(c) A low-level motion feature from PVT Stage 1 ($64\times64$ pixels; LE applied). 
(d) A high-level motion feature from PVT Stage 4 ($8\times8$ pixels; LE applied).}

\label{fig:feature map}
\end{figure}

Using PVT,
we can extract motion features embedded in a motion-blurred image, with different levels of understanding.
For example, 
Stage 1 produces local feature maps that capture fine-grained patterns with a small receptive field, detecting low-level motion blurring.
See Fig.~\ref{fig:feature map}(c) with the receptive field size of $4 \times 4$.
In Stage 4, 
we extract global feature maps, 
high-level representations that summarize information from the entire input image with a significantly large receptive field,
capturing semantic context of a motion trajectory embedded in a blurred image.
See Fig.~\ref{fig:feature map}(d) with the receptive field size of $32 \times 32$.

\subsubsection{Proposed conditioning approach: Multi-scale feature aggregation}
\label{subsubsec:feature integration}

Now, we aim to generate rich motion representations suitable for dense trajectory estimation.
Specifically, we aggregate motion features with different levels of understanding, $\mathcal{F}_\text{multi-scale}$ in (\ref{eq:feat-ms}), by using a progressive locality decoder (PLD) \cite{stepwise}.
The PLD consists of two schemes: 
local emphasis (LE) and stepwise feature aggregation (SFA). 
To effectively suppress irrelevant motion artifacts and enhance salient motion-related features,
we first apply an LE module to feature maps of each scale in $\mathcal{F}_\text{multi-scale}$ (\ref{eq:feat-ms}):
\begin{equation}
\mathbf{f}^{\text{up}}_s 
= 
\mathrm{LE}_{\boldsymbol{\zeta}_s}(\mathbf{f}_s),
\quad s = 1,\ldots,4,
\label{eq:feat-up}
\end{equation}
where \(\boldsymbol{\zeta}_s\) represents the parameters of the LE module at the \(s\)th stage.
In each LE module, we apply two single-layer convolutional networks (ConvNets) with kernels of size $3\times3$, each followed by the ReLU activation function, to \( \mathbf{f}_s \) in (\ref{eq:feat-ms}), and upsample its output to the same spatial dimension of \( H/4 \times W/4 \).
We want to extract the same motion information from every patch at each scale/PVT stage,
so we apply an LE module with the same parameters to every patch at each scale/PVT stage.
For different PVT stages, we use different LE modules.

We hypothesize that directly aggregating motion features from different PVT stages -- especially those with substantial depth discrepancies -- may lead to a motion representation gap.
To alleviate this, we use the SFA mechanism that progressively integrates motion features across different levels of the feature hierarchy, from deeper (global motion context) to shallower layers (localized motion details):
\begin{align}
\mathbf{z}_4 
&= 
\mathbf{f}_4^{\text{up}},
\nonumber
\\
\mathbf{z}_s 
&= 
\mathrm{Conv}_{\boldsymbol{\eta}_s}^{1 \times 1}
\big( 
\mathbf{z}_{s+1} \,\mycirc{\textsf{c}}\, \mathbf{f}^{\text{up}}_s 
\big), 
\quad 
s = 3,\ldots,1,
\label{eq:z1}
\end{align}
where \( \mathbf{f}_4^{\text{up}}, \ldots \mathbf{f}_1^{\text{up}} \) are given as in (\ref{eq:feat-up}), 
$\mathrm{Conv}^{1\times1}_{\boldsymbol{\eta}_s}$ denotes a single-layer ConvNet with kernels of size \(1 \times 1\) and the ReLU activation function, and parameters \(\boldsymbol{\eta}_s\) at the \(s\)th stage, 
and $\mathbf{x} \,\mycirc{\textsf{c}}\, \mathbf{x}'$ denotes concatenation of $\mathbf{x}$ and $\mathbf{x}'$ along the channel dimension.
This can be seen as gradually enriching global motion representations with fine-grained local motion cues, 
thereby minimizing representational gaps between coarse and fine motion information.

Finally, we concatenate the aggregated feature map 
\(\mathbf{z}_1\) in (\ref{eq:z1}) with the encoded features by a diffusion denoiser, by serving as a condition for the conditional diffusion model. 
The denoiser $D_{\boldsymbol{\uptheta}}$, 
a composition of an encoder $\mathrm{Enc}_{\boldsymbol{\uptheta}_\text{E}}$ and a decoder $\mathrm{Dec}_{\boldsymbol{\uptheta}_\text{D}}$,
i.e., $D_{\boldsymbol{\uptheta}} = \mathrm{Dec}_{\boldsymbol{\uptheta}_\text{D}} \circ \mathrm{Enc}_{\boldsymbol{\uptheta}_\text{E}}$, directly estimates the clean trajectory \(\hat{\mathbf{x}}_0\) from the noisy trajectory \(\mathbf{x}_t\), conditioned on the \(t\)th timestep and \(\mathbf{z}_1\) in (\ref{eq:z1}):
\begin{equation}    
\hat{\mathbf{x}}_0 
= 
\mathrm{Dec}_{\boldsymbol{\uptheta}_\text{D}}
(
\mathrm{Enc}_{\boldsymbol{\uptheta}_\text{E}}
(\mathbf{x}_t, t)
\,\mycirc{\textsf{c}}\,
\mathbf{z}_1
),
\quad 
t = T,\ldots,1,
\label{eq:denoiser}
\end{equation}
where $\boldsymbol{\uptheta}_\text{E}$ and $\boldsymbol{\uptheta}_\text{D}$ are parameters of an encoder and a decoder of $D_{\boldsymbol{\uptheta}}$, respectively.

\subsubsection{Simple diffusion denoiser architecture}

Considering the low visual complexity of motion trajectory images, 
we use a simple U-shaped network (U-Net) for $D_{\boldsymbol{\uptheta}}$.
Different from the standard diffusion U-Net architecture (see, e.g., DDPM \cite{ddpm}), 
our design is asymmetric: it has a single encoding block (for $\mathrm{Enc}_{\boldsymbol{\uptheta}_E}$) and two decoding blocks(for $\mathrm{Dec}_{\boldsymbol{\uptheta}_D}$), 
\emph{without} using skip connections.

An encoding block consists of a sequence of ConvNets (where stride is $1$ unless stated otherwise): 
single-layer ConvNet with kernels of size \(7 \times 7\), stride of $4$, and the ReLU activation function
$\rightarrow$
a ResNet block with kernels of size $3 \times 3$ and batch normalization
$\rightarrow$
single-layer ConvNet with kernels of size \(3 \times 3\) and the ReLU activation function.
The mid layer consist of a single-layer ConvNet with kernels of size \(1 \times 1\), and the ReLU activation function.
Each decoding block consists of a sequence of the following modules (where stride is $1$ for all ConvNets):
the PixelShuffle upsampling operator \cite{pixelshuffle} with an upscaling factor of $2$
$\rightarrow$
single-layer ConvNet with kernels of size \(3 \times 3\) and the ReLU activation function. 
The output layer consists of a dropout layer and a single-layer ConvNet with kernels of size \(1 \times 1\).

\subsection{Proposed loss function}
\label{sec:loss}

Our target, an HR motion trajectory map, is a 2D binary image of the same spatial dimensions as the input blurred image. 
It is supposed to consist of connected trajectory pixels with a value of $1$ and background pixels with a value of $0$.
We use a combination of binary cross-entropy (BCE) and intersection-over-union (IoU) loss between the ground-truth and estimated HR motion trajectory maps.
The BCE loss promotes accurate classification of each pixel as either motion trajectory or background, 
and the IoU loss promotes consistent estimation of the global structure and spatial alignment of a motion trajectory, by maximizing the spatial overlap between estimated and ground-truth trajectories.

To increase the importance of trajectory pixels relative to background pixels in BCE and IoU losses, we define the pixel-wise weights $\mathbf{w} \in \mathbb{R}^{H \times W}$:
\begin{equation}
\label{eq:weight}
\mathbf{w} = \lambda \cdot \mathbf{x}_0 + \mathbf{1}, 
\end{equation}
where $\lambda \in \mathbb{R}_{>0}$ is a hyperparameter and note that the ground truth motion trajectory $\mathbf{x}_0 \in \{ 0, 1 \}^{H \times W}$.
This can also consider that in general, an HR motion trajectory map is sparse.

Finally, we propose the loss function for training MoTDiff as follows:
\begin{equation}
\label{eq:overall loss}
\mathcal{L}_{\text{MoTDiff}} = \mathcal{L}_{\text{wBCE}}(\hat{\mathbf{x}}_0, \mathbf{x}_0) + \mathcal{L}_{\text{wIoU}}(\hat{\mathbf{x}}_0, \mathbf{x}_0),
\end{equation}
where $\mathcal{L}_{\text{wBCE}}$ and $ \mathcal{L}_{\text{wIoU}}$ are the weighted BCE loss and the weighted IoU loss \cite{f3net} using the pixel-wise weights $\mathbf{w}$ in (\ref{eq:weight}), respectively. 
We train the proposed MoTDiff in an end-to-end manner by minimizing the loss function $\mathcal{L}_{\text{MoTDiff}}$ in (\ref{eq:overall loss}) with respect to all the parameters of MoTDiff, $\{ \boldsymbol{\xi}, \boldsymbol{\zeta}_s, \boldsymbol{\eta}_s, \boldsymbol{\uptheta} : s = 1,\ldots,4 \}$.
In training the MoTDiff, to prevent overfitting to specific steps, 
we uniformly randomly sample a timestep $t \in \{1, \ldots, T\}$ at each iteration, similarly in \cite{ddpm}.

\subsection{Proposed stochastic trajectory pixel dropout (STPD)}
\label{sec:STPD}

Trained MoTDiff by minimizing the proposed loss (\ref{eq:overall loss}) can identify trajectory pixels while preserving the overall structure of target motion paths of a camera.
However, minimizing (\ref{eq:overall loss}) alone may be insufficient to promote the connectivity between points in a generated motion trajectory,
giving a fragmented/disconnected motion path.

To resolve this drawback, 
we propose a new STPD strategy that can encourage the proposed MoTDiff to generate a spatially connected motion trajectory.
In the foraward process of MoTDiff,
we modify (\ref{eq:forward_process}) as follows:
\begin{align}
\label{eq:stpd}
\mathbf{x}_0' 
&= \mathrm{STPD}(\mathbf{x}_0,\, p);
\\
\mathbf{x}_t' 
&= \sqrt{\alpha_t}\,\mathbf{x}_0' + \sqrt{1 -\alpha_t}\,\boldsymbol{\epsilon}, 
\quad t=1,\ldots,T,
\nonumber
\end{align}
where $\mathrm{STPD}(\mathbf{x}_0,\, p)$ is the proposed STPD strategy that randomly changes trajectory pixels in the ground truth trajectory map $\mathbf{x}_0$ to background with probability $p \in (0,1)$.
By introducing intentional disconnections, 
we can encourage our MoTDiff to recover fragmented trajectory estimates and reinforce the spatial connectivity of motion trajectories.

\section{Experimental results and discussion}
\label{sec:exp}

This section provides details of experimental setups and results with some discussion. 
We evaluated MoTDiff for two compuatational imaging tasks, blind image deblurring and CEP, with particular emphasis on spatially invariant motion.
In blind image deblurring, 
we compared proposed HR trajectory-based MoTDiff with the PSF-based SOTA methods, Kernel-Diff \cite{kernel-diff}, BlindDPS \cite{blind-dps}, SelfDeblur \cite{selfdeblur}, and PMP \cite{pmp}, and the parametric trajectory-based SOTA method, Motion-ETR \cite{ETR}.
In CEP, we compared proposed HR trajectory-based MoTDiff with the following SOTA methods: 
the blur length-based method \cite{dnfcep}, frame-based method \cite{dce}, and parametric trajectory-based method, Motion-ETR \cite{ETR}.

\subsection{Experimental setups}

\subsubsection{Datasets}
\label{sec:data}

We constructed synthetic datasets by blurring sharp images with simulated PSFs -- that are resampled from simulated HR motion trajectories -- similar to the standard image deblurring experimental setup \cite{blind-dps, kernel-diff}.
We randomly selected sharp images from the GoPro dataset \cite{gopro}, and for each image, we extracted a randomly cropped patch of size $256 \times 256$.
In image blurring simulation, we used the symmetric boundary condition, following \cite{kernel-diff}.
We generated random HR motion trajectories and their corresponding PSFs with size of $64 \times 64$, following the simulation pipeline in \cite{trajmodeling}.
For training, 
we constructed a synthetic dataset by simulating 50k blurred images using 50 sharp images from the GoPro train dataset and simulated 1k PSFs and HR motion trajectories.
To test trained models, 
we constructed a synthetic dataset by simulating 1k blurred images using 10 sharp images from the GoPro test dataset and simulated 100 PSFs (of size $64 \times 64$) and HR motion trajectories (of size $256 \times 256$).

In evaluating different blind image deblurring models, 
we used two datasets.
The first test dataset is the synthetic dataset simulated using the GoPro data; see above. 
To evaluate real-world blind image deblurring performances of trained models,
we randomly selected 100 real blurred images from the RSBlur dataset \cite{rsblur}.
Noting that the focus of this work is spatially invariant blur, we randomly cropped a patch of size $256\times256$ from each motion-blurred image in the real world, 
following the setting in \cite{kernel-diff}.
In evaluating different CEP methods,
we used the synthetic GoPro test dataset above.
We could not run CEP experiments with the real-world datasets as they do \emph{not} have ground-truth motion trajectories to mimic CEP.

\subsubsection{Experimental setups for blind image deblurring}
\label{sec:debcep}

We first describe the blind image deblurring setup of proposed MoTDiff.
In deblurring motion in an observed image via proposed MoTDiff,
we first obtained a PSF of size $64 \times 64$ by resampling an estimated HR motion trajectory using the sub-pixel linear interpolation method \cite{trajmodeling}.
We then used the iterative non-blind image deblurring optimization method \cite{l0prior} using a given PSF above, following the setup in \cite{pmp}.

Now, we describe the blind image deblurring setup for Motion-ETR \cite{ETR}, a spatially variant motion estimation method that estimates a parametric trajectory for each pixel. 
We obtained a blur kernel, a.k.a., motion PSF, of size $64 \times 64$ for each pixel with a procedure similar to the one used in MoTDiff experiments.
For fair comparisons with the remaining blind deblurring methods for spatially invariant motion blur, 
we adapted the iterative non-blind deblurring method \cite{l0prior} used above, after identifying a single representative PSF.
We obtained a single representative PSF by computing the similarity between the PSF at each pixel and those at other pixel locations, and choosing the one with the highest average similarity; we refer to this as an oracle spatially invariant setup.
For computational efficiency, we considered PSFs within the central region of the image with size of $100\times100$.

For the remaining SOTA blind image deblurring methods (see Section~\ref{sec:exp}), we used their default setups.
By default, the blur kernel size is set to $64 \times 64$.

\subsubsection{Experimental setups for CEP}

We conducted CEP experiments under the standard CEP assumption of consistent motion between the initial calibration imaging 
for estimating a motion trajectory and/or optimizing a code, i.e., shutter fluttering pattern,
and subsequent CEP by modulating motion with an optimized code.

In estimating a motion trajectory and optimizing a code for proposed MoTDiff and Motion-ETR \cite{ETR}, 
we used the synthesized GoPro test dataset (see Section~\ref{sec:data}) that mimics initial calibration imaging.
In optimizing codes for the proposed MoTDiff and Motion-ETR \cite{ETR}, 
we used an estimated HR trajectory and parametric trajectory, respectively.
In optimizing codes by DNF \cite{dnfcep}, we used the trajectory length that is assumed to be known.
In optimizing codes by DCE \cite{dce}, we used video frames that were simulated by shifting a sharp image along the downsampled simulated HR trajectory, where we used the GoPro test dataset (see Section~\ref{sec:data}).

To mimic CEP using an optimized code under the assumption of the same camera motion as during the initial exposure,
we blurred sharp images -- that were used in constructing the synthetic test dataset in Section~\ref{sec:data} -- with coded PSFs of size $64 \times 64$.
We generated a coded PSF by modulating the corresponding HR ground-truth trajectory with an optimized code and resampling a modulated result.

We investigated the effectiveness of different CEP methods in motion deblurring using coded PSFs,
by using simple inverse filtering using the coded PSF above rather than advanced image deblurring algorithms, 
following the setup in \cite{firstCEP}.
In addition, 
we investigated the invertibility of CEP using coded PSFs by visualizing its modulation transfer function (MTF) \cite{firstCEP}.

\subsubsection{Implementation details}

We first provide implementation details of the proposed MoTDiff framework.
In extracting multi-scale features from a blurred image, 
we used the PVT v2 backbone \cite{pvtv2},
where we initialized its weights with pretrained ones using the ImageNet dataset \cite{imagenet}.
We trained our MoTDiff for 60k iterations on a single NVIDIA A100 GPU with a batch size of 128. 
We used the Adam optimizer with an initial learning rate of $1 \times 10^{-4}$, applying the cosine annealing schedule to gradually reduce the learning rate to a minimum of $1 \times 10^{-6}$. 
We set the diffusion timestep \( T = 1000 \) and used the same number of steps for sampling, using the cosine noise schedule \cite{iddpm}.

For the current SOTA methods listed at the beginning of Section~\ref{sec:exp}, we used the default configurations specified in their respective papers.

\subsubsection{Evaluation metrics}

As evaluation metrics, we used the maximum of normalized cross-correlation (MNC) \cite{mnc}, the peak signal-to-noise ratio (PSNR), and the structural similarity index measure (SSIM) \cite{ssim}. 
The MNC metric evaluates the accuracy of the estimated PSFs, where we note again that we resampled estimated HR motion trajectories from proposed MoTDiff to generate PSFs.
The PSNR and SSIM metrics evaluate the quality of deblurred images.

\subsection{Comparisons between different blind image deblurring models}

\begin{figure*}[t!] 
\centering
\setlength{\tabcolsep}{0.1pt}
\begin{tabular}{cccccccc}
     \includegraphics[width=0.125\textwidth]{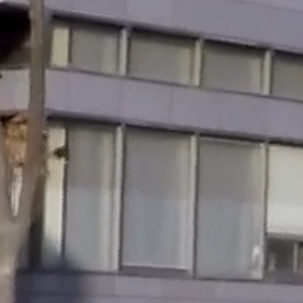} & \includegraphics[width=0.125\textwidth]{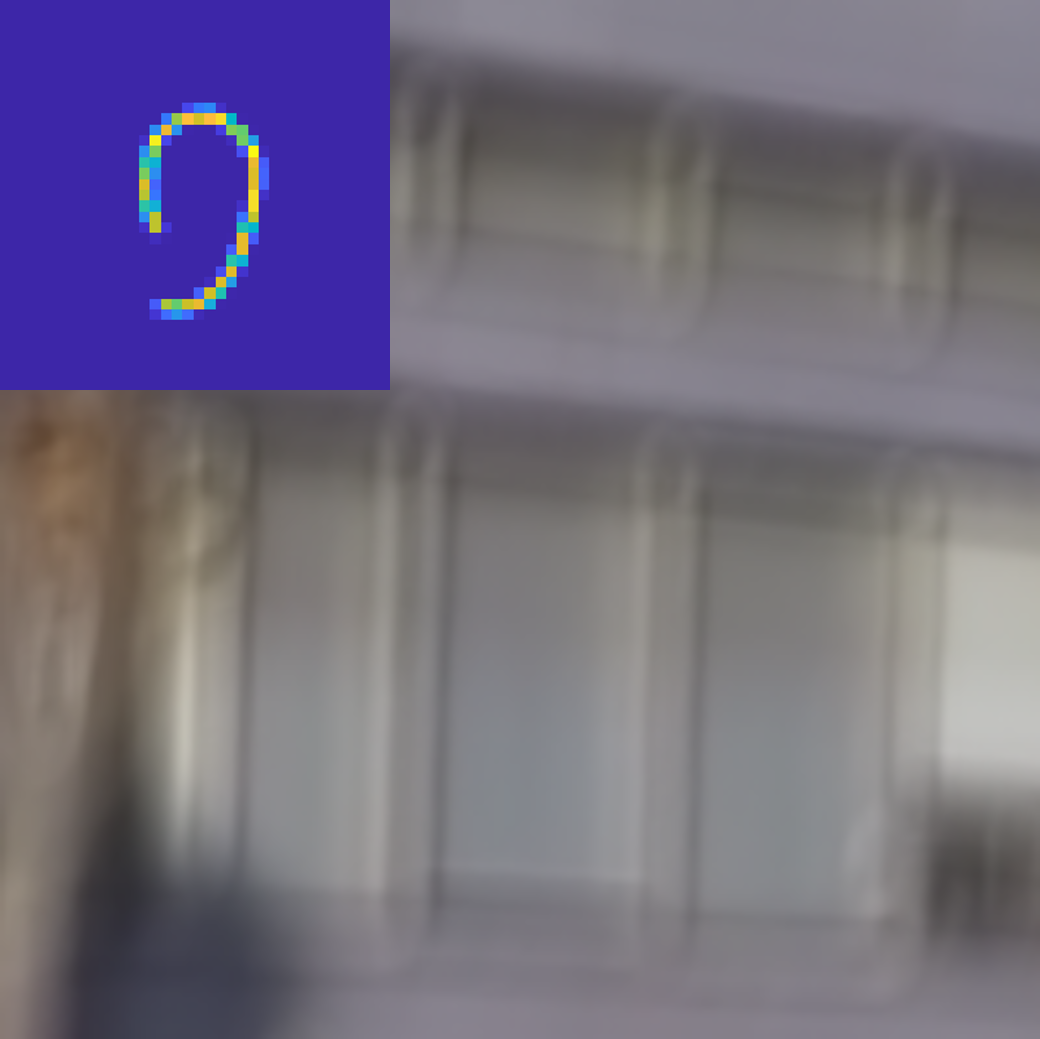} & \includegraphics[width=0.125\textwidth]{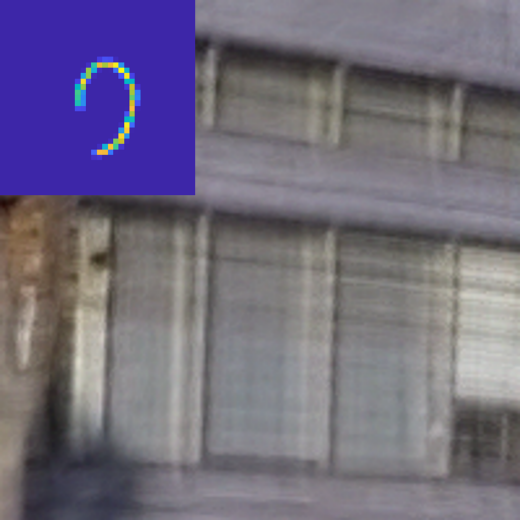} & \includegraphics[width=0.125\textwidth]{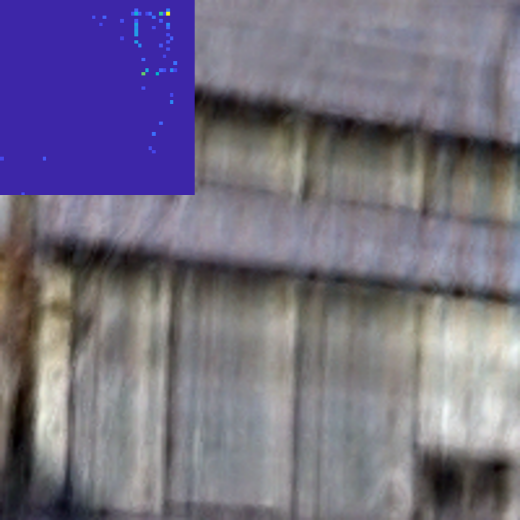} & \includegraphics[width=0.125\textwidth]{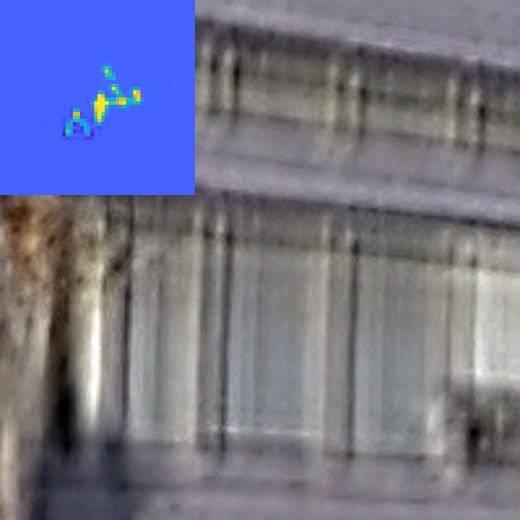} & \includegraphics[width=0.125\textwidth]{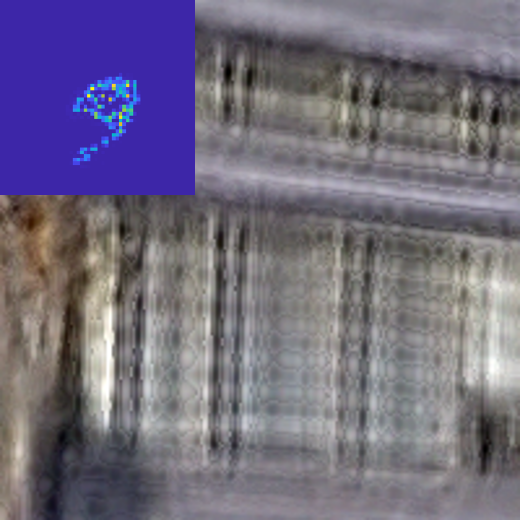} & \includegraphics[width=0.125\textwidth]{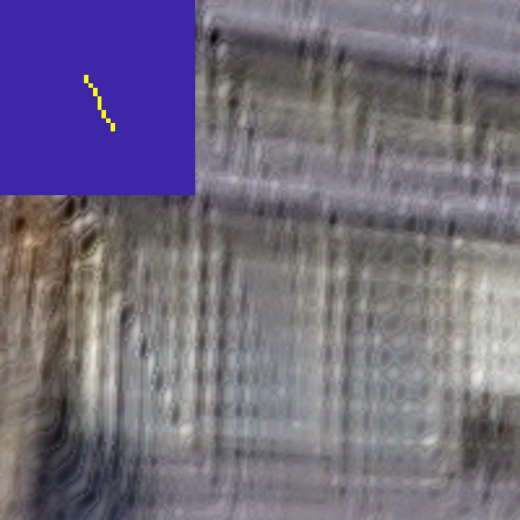} & \includegraphics[width=0.125\textwidth]{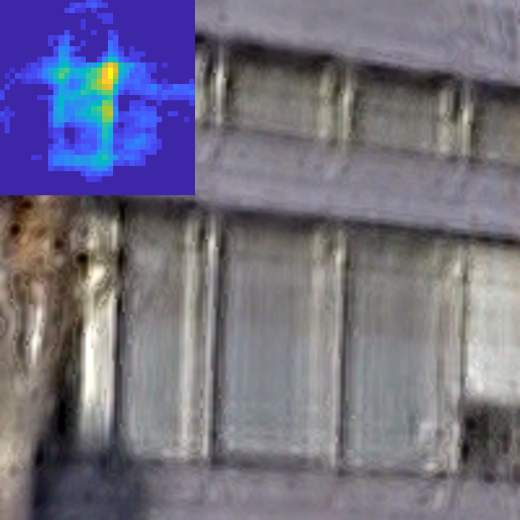}    \\ 
     \includegraphics[width=0.125\textwidth]{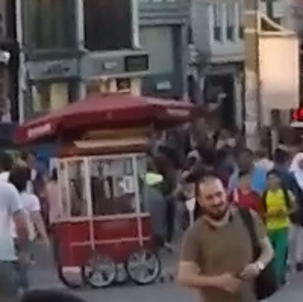} & \includegraphics[width=0.125\textwidth]{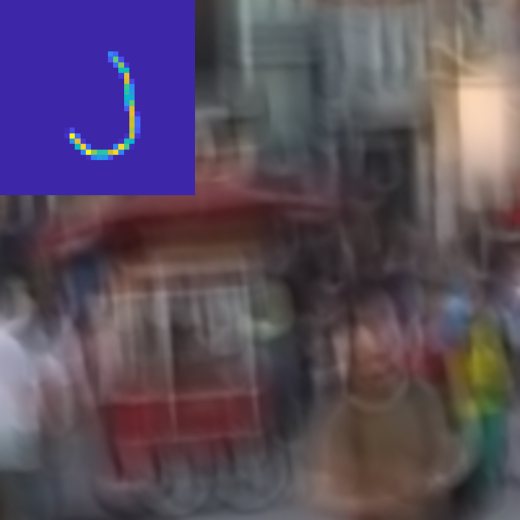} & \includegraphics[width=0.125\textwidth]{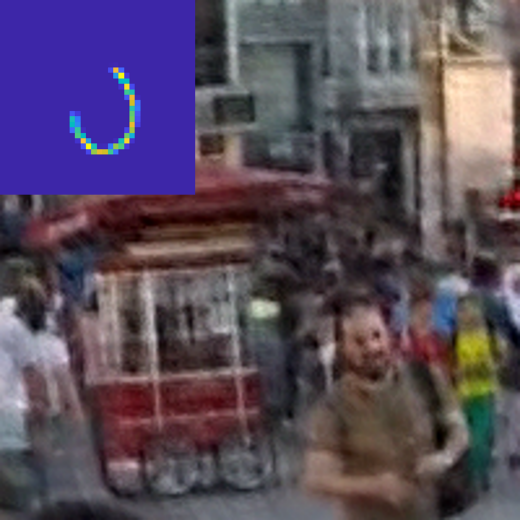} & \includegraphics[width=0.125\textwidth]{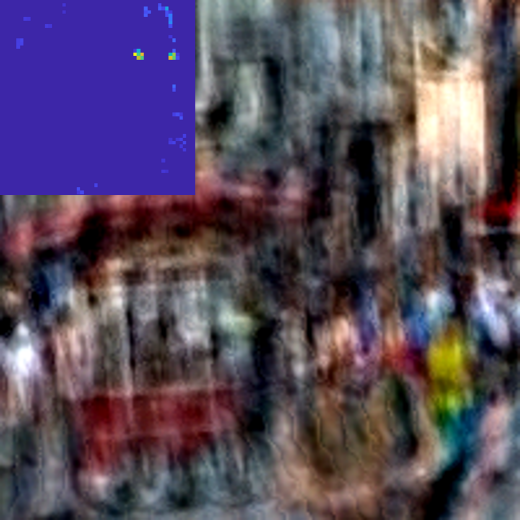} & \includegraphics[width=0.125\textwidth]{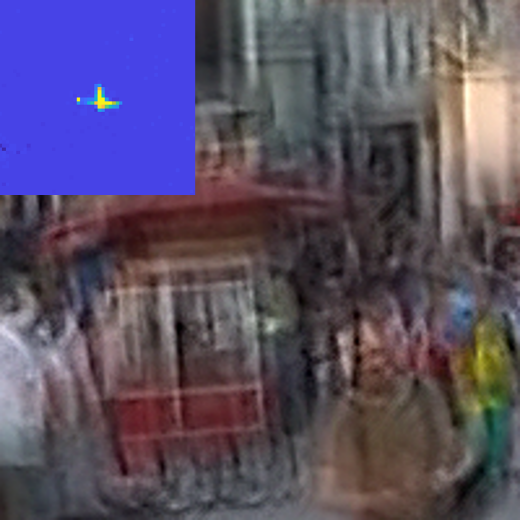} & \includegraphics[width=0.125\textwidth]{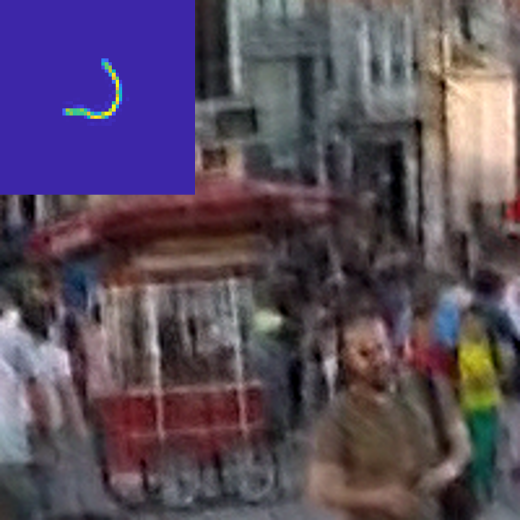} & \includegraphics[width=0.125\textwidth]{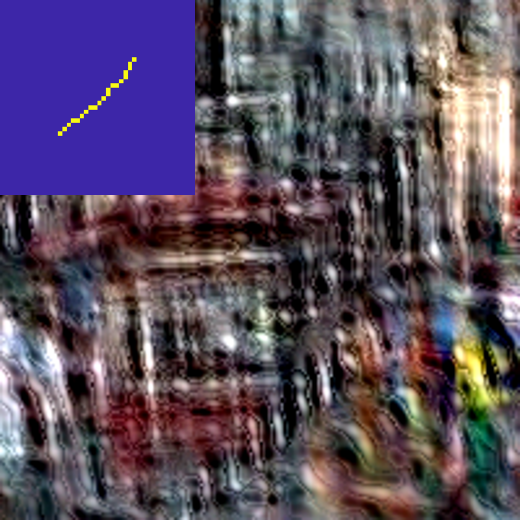} & \includegraphics[width=0.125\textwidth]{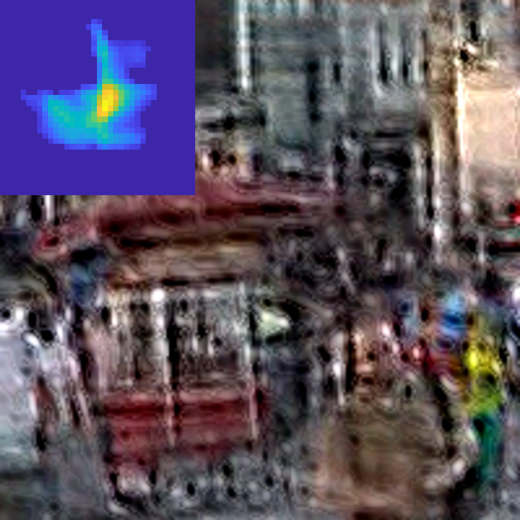}    \\ 
     \includegraphics[width=0.125\textwidth]{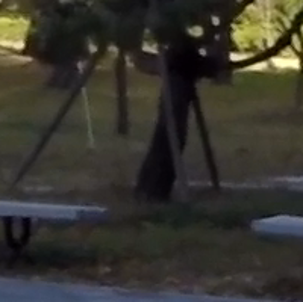} & \includegraphics[width=0.125\textwidth]{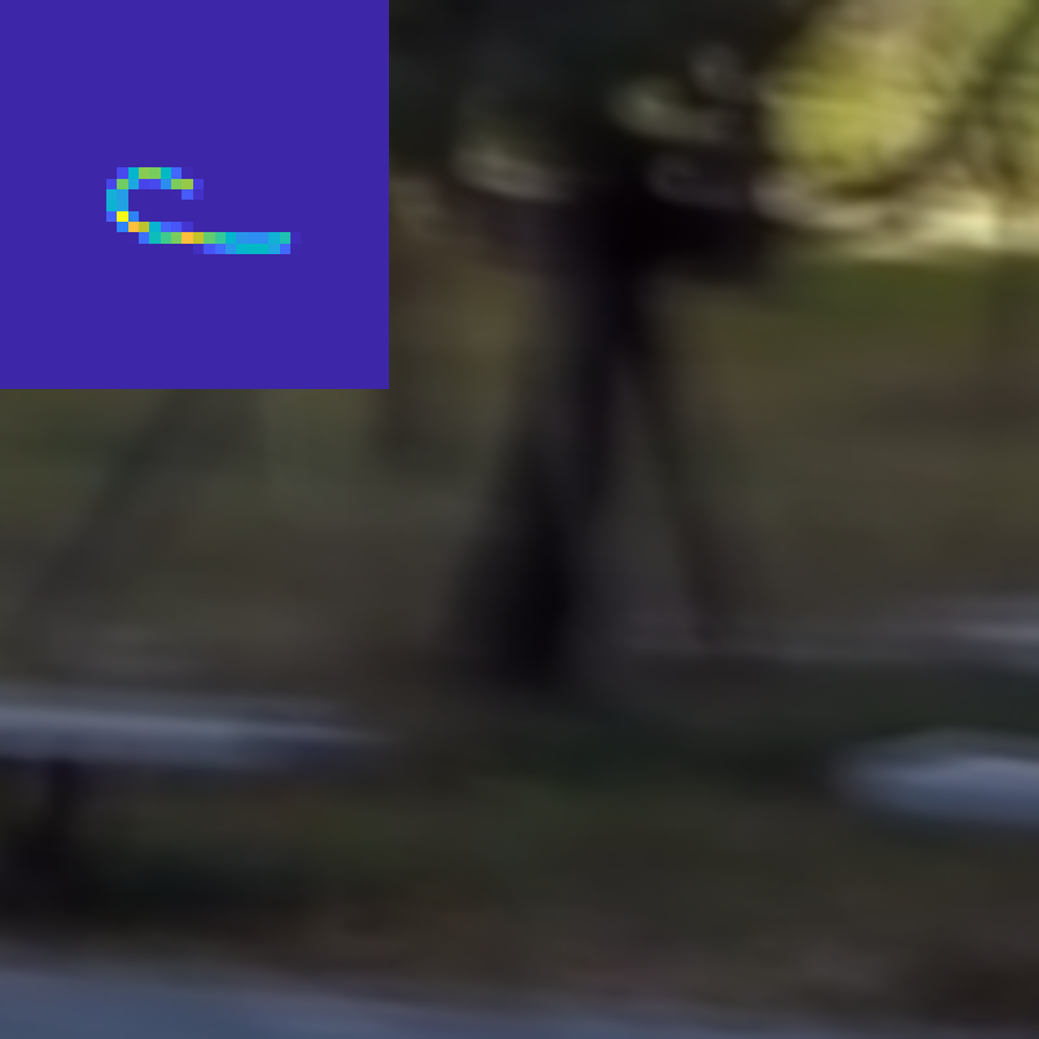} & \includegraphics[width=0.125\textwidth]{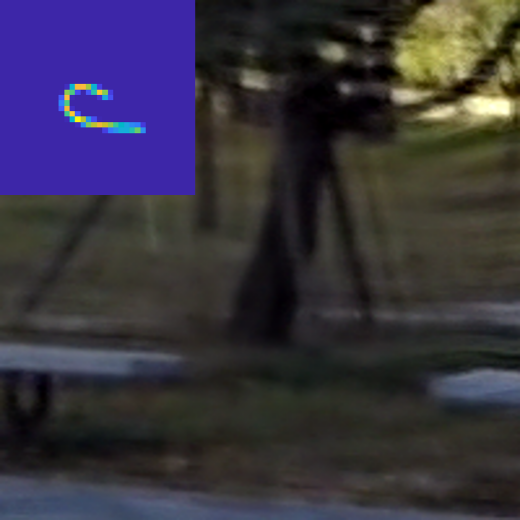} & \includegraphics[width=0.125\textwidth]{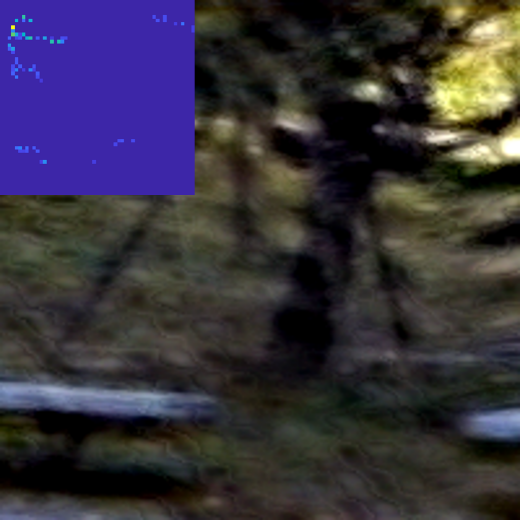} & \includegraphics[width=0.125\textwidth]{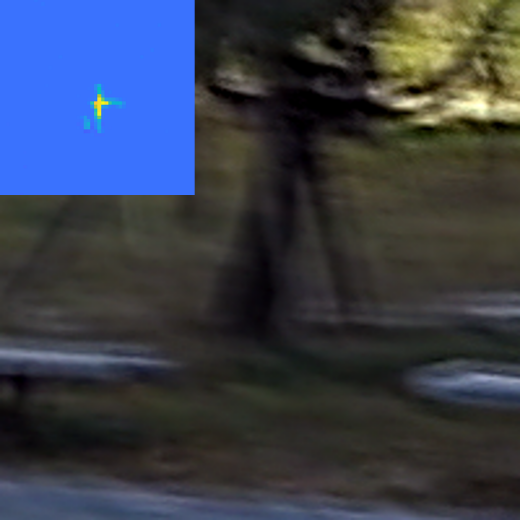} & \includegraphics[width=0.125\textwidth]{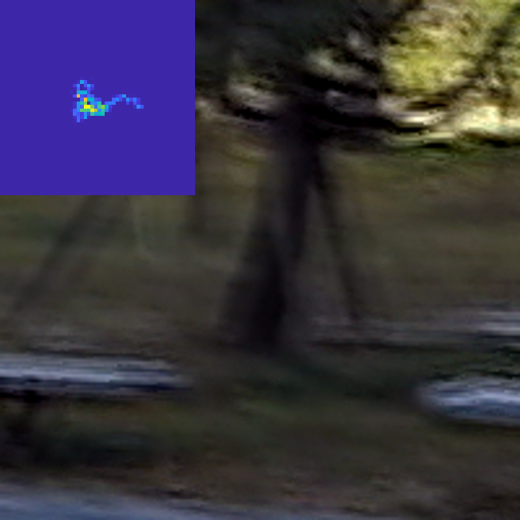} & \includegraphics[width=0.125\textwidth]{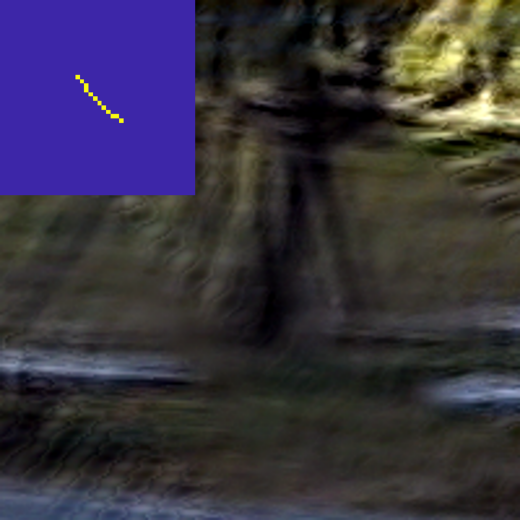} & \includegraphics[width=0.125\textwidth]{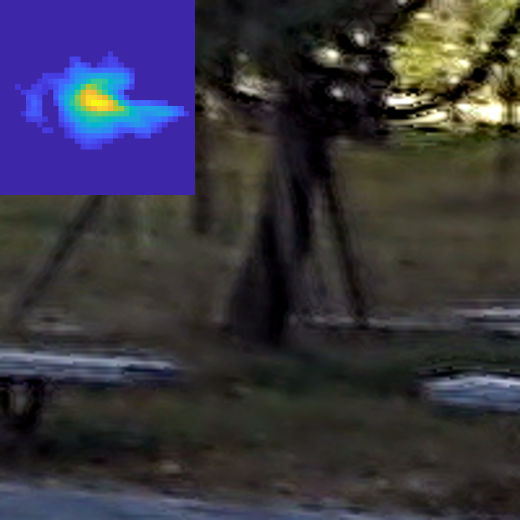} \\
     \includegraphics[width=0.125\textwidth]{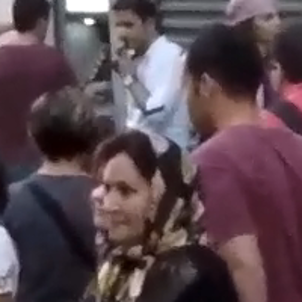} &  \includegraphics[width=0.125\textwidth]{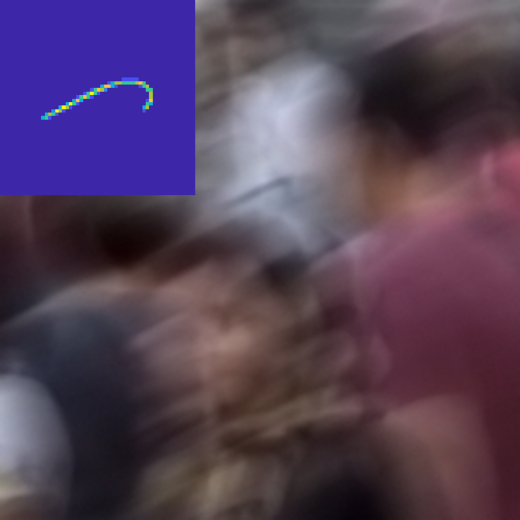} & \includegraphics[width=0.125\textwidth]{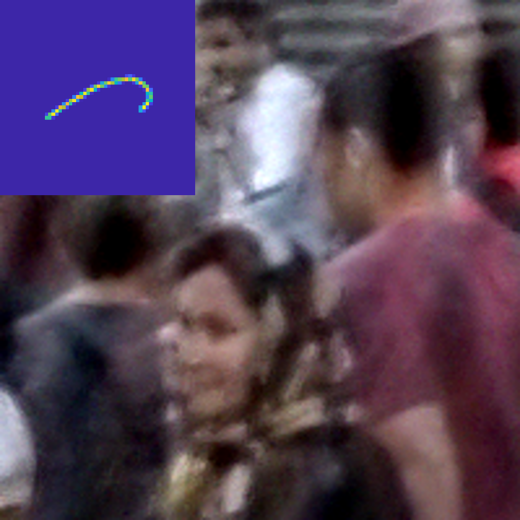} & \includegraphics[width=0.125\textwidth]{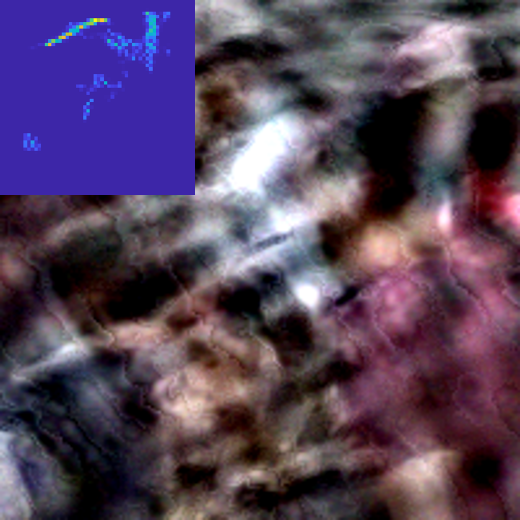} & \includegraphics[width=0.125\textwidth]{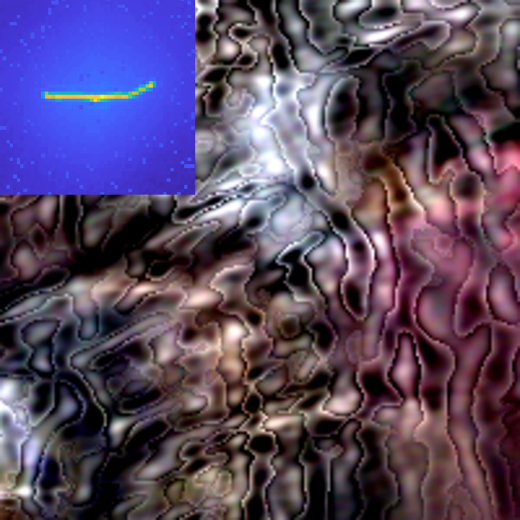} & \includegraphics[width=0.125\textwidth]{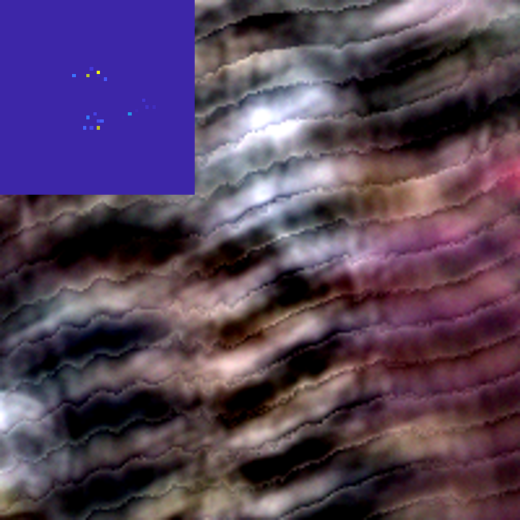} & \includegraphics[width=0.125\textwidth]{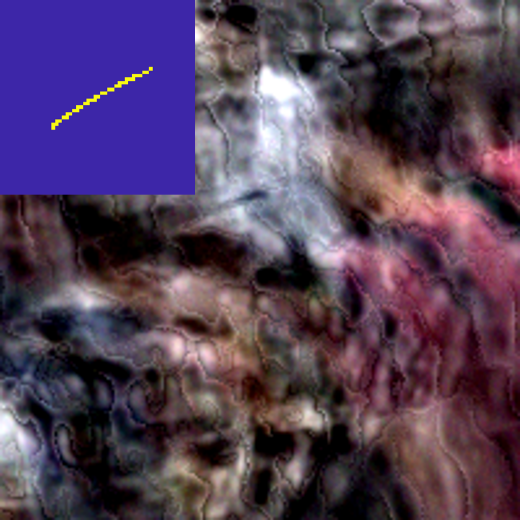} & \includegraphics[width=0.125\textwidth]{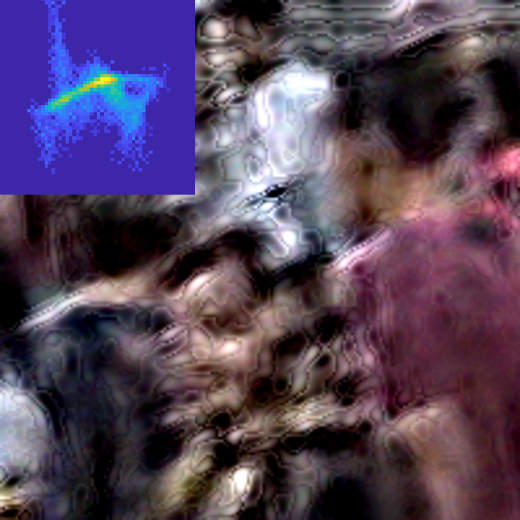} \\
     
     \parbox[t]{0.125\textwidth}{\centering\footnotesize Ground-truth\label{fig:fig3(a)}} &
     \parbox[t]{0.125\textwidth}{\centering\footnotesize Blurred image\label{fig:fig3(b)}} &
     \parbox[t]{0.125\textwidth}{\centering\footnotesize \textbf{MoTDiff {\scriptsize(ours)}}\label{fig:fig3(c)}} &
     \parbox[t]{0.125\textwidth}{\centering\footnotesize  SelfDeblur \cite{selfdeblur}\label{fig:fig3(d)}} &
     \parbox[t]{0.125\textwidth}{\centering\footnotesize BlindDPS \cite{blind-dps}\label{fig:fig3(e)}} &
     \parbox[t]{0.125\textwidth}{\centering\footnotesize Kernel-Diff \cite{kernel-diff}\label{fig:fig3(f)}} &
     \parbox[t]{0.125\textwidth}{\centering\footnotesize Motion-ETR \cite{ETR}\label{fig:fig3(g)}} &
     \parbox[t]{0.125\textwidth}{\centering\footnotesize PMP \cite{pmp}\label{fig:fig3(h)}}
     
\end{tabular}

\caption{Comparisons of deblurred images and estimated PSFs from different blind image deblurring methods (the inset in the top-left corner displays ground truth or estimated PSF; we used the synthetic GoPro dataset in Section~\ref{sec:data}). 
The proposed MoTDiff can give significantly better motion trajectories and deblurred images compared to the several SOTA blind image deblurring methods.}
\label{fig:psf-deblur-gopro}
\end{figure*}

\begin{figure*}[t!] 
\centering
\setlength{\tabcolsep}{0.1pt}
\begin{tabular}{cccccccc}
\includegraphics[width=0.125\textwidth]{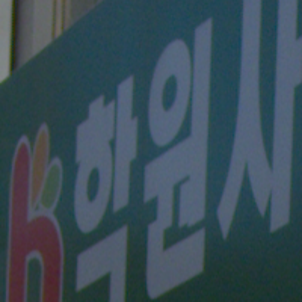} &
\includegraphics[width=0.125\textwidth]{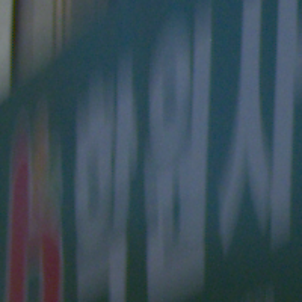} &
\includegraphics[width=0.125\textwidth]{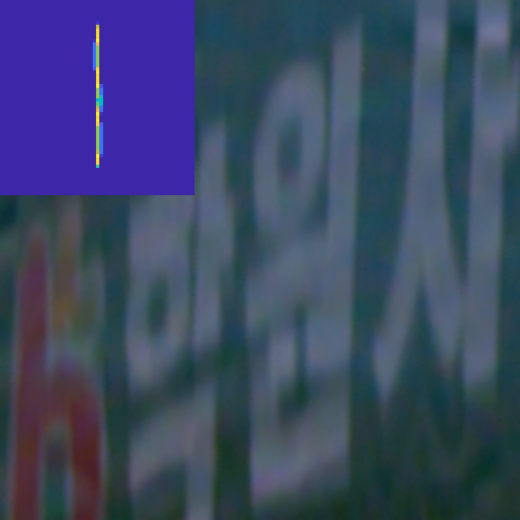} &
\includegraphics[width=0.125\textwidth]{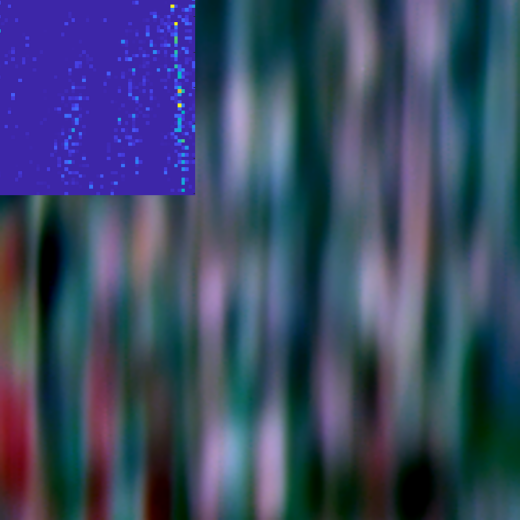} &
\includegraphics[width=0.125\textwidth]{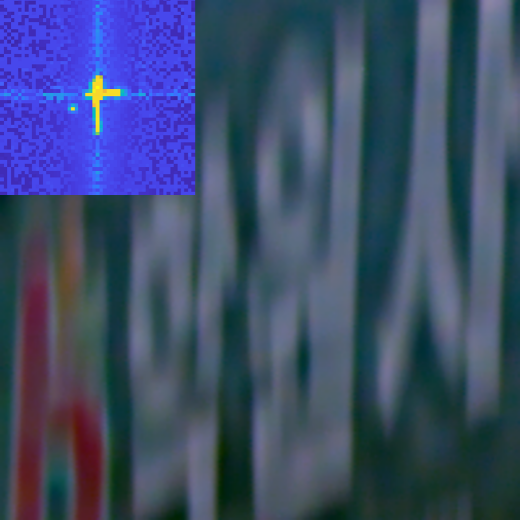} &
\includegraphics[width=0.125\textwidth]{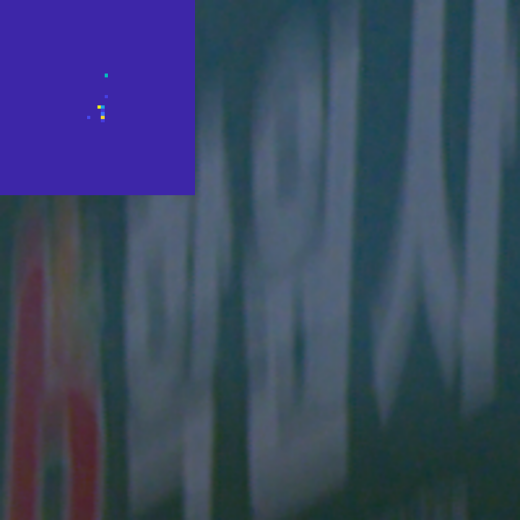} &
\includegraphics[width=0.125\textwidth]{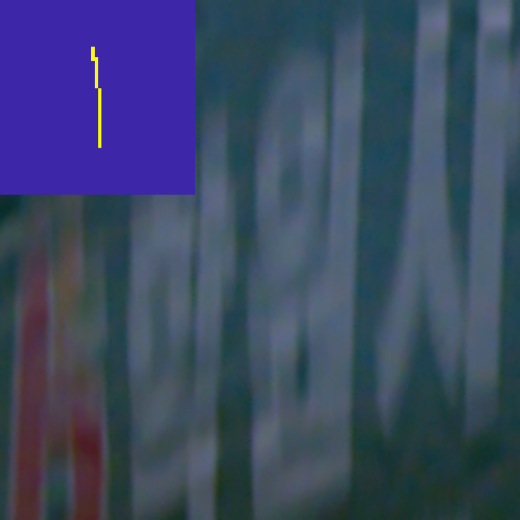} &
\includegraphics[width=0.125\textwidth]{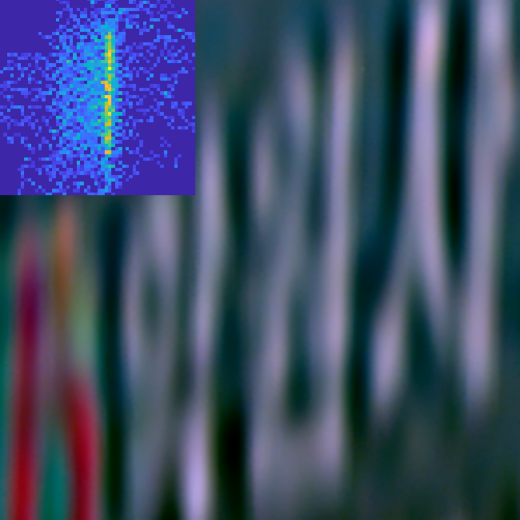} \\

\includegraphics[width=0.125\textwidth]{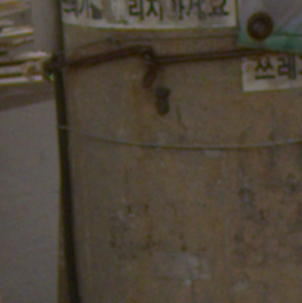} &
\includegraphics[width=0.125\textwidth]{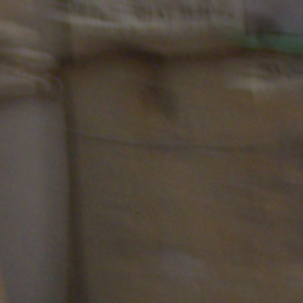} &
\includegraphics[width=0.125\textwidth]{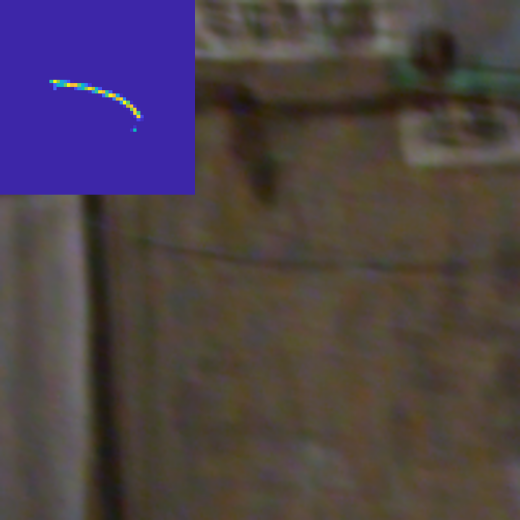} &
\includegraphics[width=0.125\textwidth]{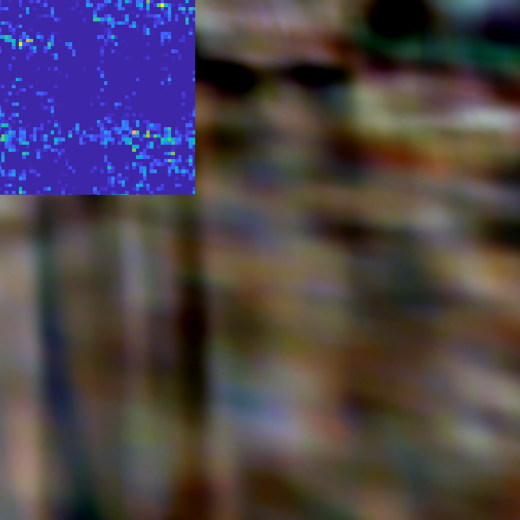} &
\includegraphics[width=0.125\textwidth]{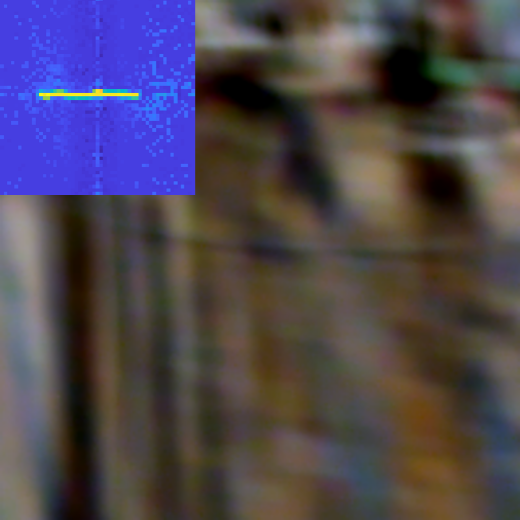} &
\includegraphics[width=0.125\textwidth]{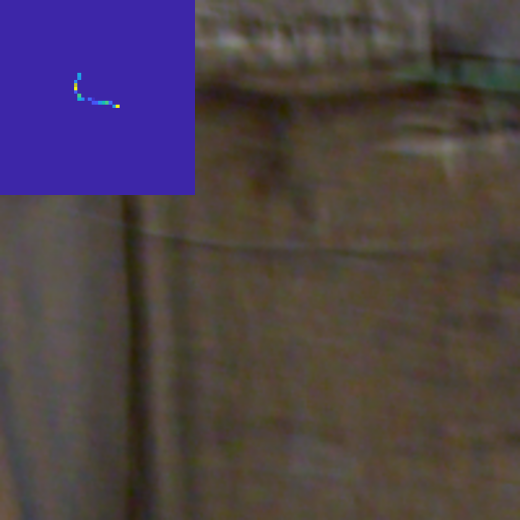} &
\includegraphics[width=0.125\textwidth]{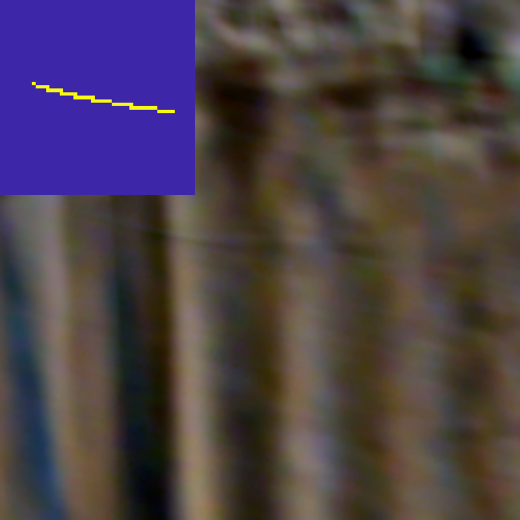} &
\includegraphics[width=0.125\textwidth]{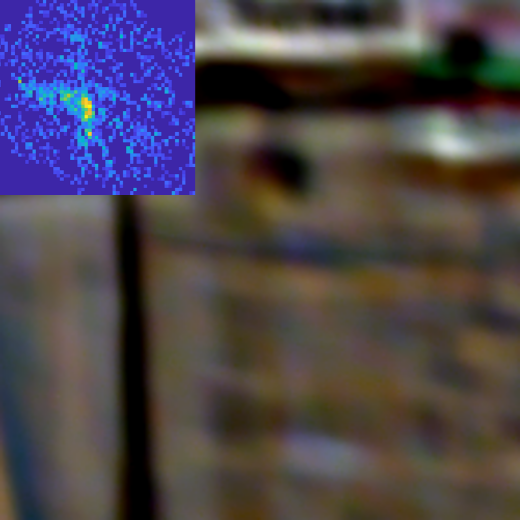} \\

\parbox[t]{0.125\textwidth}{\centering\footnotesize Ground-truth} &
\parbox[t]{0.125\textwidth}{\centering\footnotesize Blurred image} &
\parbox[t]{0.125\textwidth}{\centering\footnotesize \textbf{MoTDiff (ours)}} &
\parbox[t]{0.125\textwidth}{\centering\footnotesize SelfDeblur \cite{selfdeblur}} &
\parbox[t]{0.125\textwidth}{\centering\footnotesize BlindDPS \cite{blind-dps}} &
\parbox[t]{0.125\textwidth}{\centering\footnotesize Kernel-Diff \cite{kernel-diff}} &
\parbox[t]{0.125\textwidth}{\centering\footnotesize Motion-ETR \cite{ETR}} &
\parbox[t]{0.125\textwidth}{\centering\footnotesize PMP \cite{pmp}}

\end{tabular}

\caption{Comparisons of deblurred images and estimated PSFs from different blind image deblurring methods (the inset in the top-left corner
displays estimated PSF; we used the \emph{real-world} RSBlur dataset in Section~\ref{sec:data}).}
\label{fig:psf-deblur-rs}
\end{figure*}

\begin{table}[t!]
\caption{Performance comparisons between different blind image deblurring or motion estimation methods (synthetic GoPro test dataset).}
\label{tab:psf-deblur-gopro}
\centering\small
\begin{tabular}{lcccc}
\hline\hline
\multirow{2}{*}{Methods} & (PSF est.) & \multicolumn{2}{c}{(Blind deblurring)}
\\
& MNC $\uparrow$ & PSNR $\uparrow$ & SSIM $\uparrow$ 
\\ \hline
PMP\cite{pmp} & 0.43 & 16.89 & 0.50 
\\
SelfDeblur\cite{selfdeblur} & 0.47 & 13.05 & 0.34 
\\
BlindDPS \cite{blind-dps} & 0.32 & 13.70 & 0.34 
\\
Kernel-Diff \cite{kernel-diff} & 0.28 & 19.33 & 0.63 
\\
Motion-ETR \cite{ETR} & 0.26 & 18.94 & 0.58 
\\
\textbf{MoTDiff (ours)}  & \textbf{0.76} & \textbf{23.89} & \textbf{0.77} 
\\
\hline\hline
\end{tabular}
\end{table}

\begin{table}[t!]
\caption{Performance comparisons between different blind image deblurring or motion estimation methods (\emph{real-world} RSBlur dataset \cite{rsblur} that does not have ground-truth motion trajectories).}
\label{tab:psf-deblur-rs}
\centering\small

\begin{tabular}{lcccc}
\hline\hline
\multirow{2}{*}{Methods} & 
\multicolumn{2}{c}{(Blind deblurring)}
\\
& PSNR $\uparrow$ & SSIM $\uparrow$ 
\\ \hline
PMP\cite{pmp} & 19.14 & 0.44 
\\ 
SelfDeblur\cite{selfdeblur} & 14.79 & 0.29 
\\
BlindDPS \cite{blind-dps} & 20.16 & 0.49 
\\
Kernel-Diff \cite{kernel-diff} & 21.73 & 0.56
\\
Motion-ETR \cite{ETR} & 20.07 & 0.56 
\\
\textbf{MoTDiff (ours)}  & \textbf{22.08} & \textbf{0.67} 
\\
\hline\hline
\end{tabular}
\end{table}

The PSF estimation results in Fig.~\ref{fig:psf-deblur-gopro} and Table~\ref{tab:psf-deblur-gopro} show that the proposed MoTDiff outperforms the existing SOTA blind image deblurring or motion estimation methods from the perspective of motion estimation.
The results suggest that accurately estimating high-resolution motion trajectories leads to improved PSF estimation, ultimately improving deblurring performances.

The blind image deblurring results in Figures~\ref{fig:psf-deblur-gopro}--\ref{fig:psf-deblur-rs} and Tables~\ref{tab:psf-deblur-gopro}--\ref{tab:psf-deblur-rs} demonstrate that the proposed MoTDiff can achieve significantly better image deblurring performances compared to the existing SOTA blind image deblurring or motion estimation methods, on both the synthetic and real-world datasets.

Particularly in comparison with Motion-ETR that can estimate PSFs for different locations,
proposed MoTDiff achieved significantly better performance in spatially invariant deblurring.
Motion-ETR is fundamentally limited in capturing complex motion, because it uses a parametric trajectory representation (see Section~\ref{sec:HR motion trajectory}).
This can be observed estimated PSFs in Fig.~\ref{fig:psf-deblur-gopro} (Motion-ETR), even with the oracle spatially invariant setup in Section~\ref{sec:debcep}.

\begin{figure}[t!] 
\centering
\setlength{\tabcolsep}{0.1pt}
\begin{tabular}{ccccc}
\includegraphics[width=0.2\columnwidth]{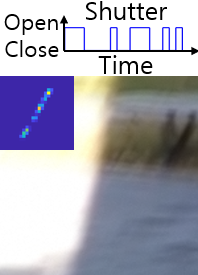} &
\includegraphics[width=0.2\columnwidth]{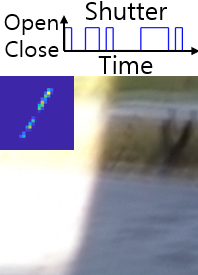} &
\includegraphics[width=0.2\columnwidth]{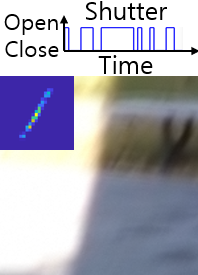} &
\includegraphics[width=0.2\columnwidth]{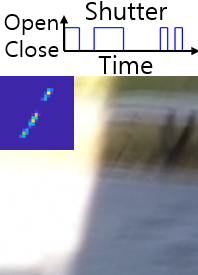} &
\includegraphics[width=0.2\columnwidth]{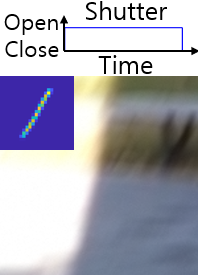} \\
\includegraphics[width=0.2\columnwidth]{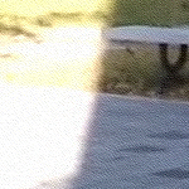} &
\includegraphics[width=0.2\columnwidth]{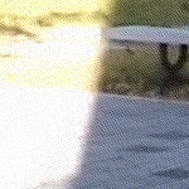} &
\includegraphics[width=0.2\columnwidth]{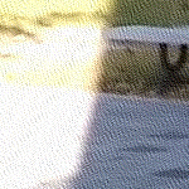} &
\includegraphics[width=0.2\columnwidth]{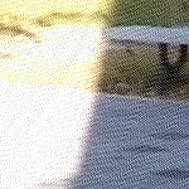} &
\includegraphics[width=0.2\columnwidth]{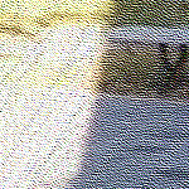} \\
\includegraphics[width=0.2\columnwidth]{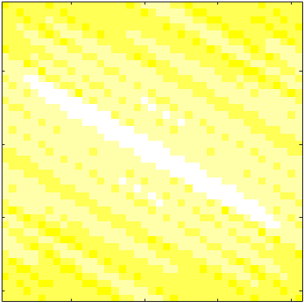} &
\includegraphics[width=0.2\columnwidth]{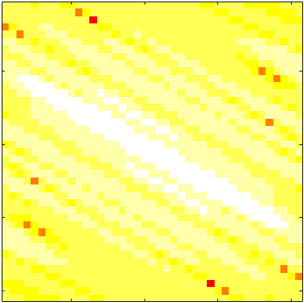} &
\includegraphics[width=0.2\columnwidth]{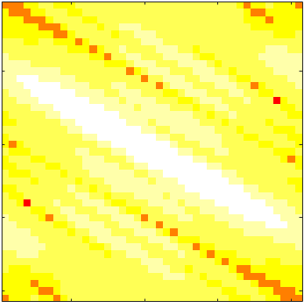} &
\includegraphics[width=0.2\columnwidth]{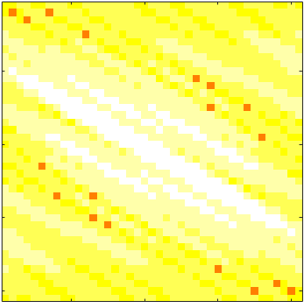} &
\includegraphics[width=0.2\columnwidth]{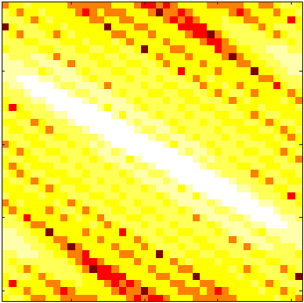} \\
\multicolumn{2}{c}{\includegraphics[width=0.4\columnwidth]{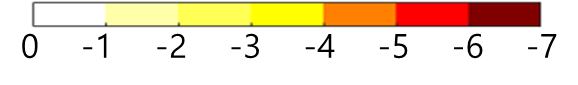}} \vspace{-0.5pc}
\\

\parbox[t]{0.2\columnwidth}{\centering\footnotesize \textbf{MoTDiff (ours)}\label{fig:fig5(a)}} &
\parbox[t]{0.2\columnwidth}{\centering\footnotesize Motion-ETR \cite{ETR}\label{fig:fig5(b)}} &
\parbox[t]{0.2\columnwidth}{\centering\footnotesize DCE \cite{dce}\label{fig:fig5(c)}} &
\parbox[t]{0.2\columnwidth}{\centering\footnotesize DNF \cite{dnfcep}\label{fig:fig5(d)}} &
\parbox[t!]{0.2\columnwidth}{\centering\footnotesize W/o CEP\label{fig:fig5(e)}} 
\end{tabular}

\caption{Comparisons of deblurred images and MTFs from different CEP methods (synthetic GoPro test dataset using optimized codes). 
Top row: Blurred images with optimized codes (corresponding optimized codes are displayed above each image; the inset in the top-left corner displays coded PSF). 
Middle row: Deblurred images using coded PSFs. 
Bottom row: MTFs of coded PSFs.}
\label{fig:CEP}
\end{figure}

\subsection{Comparisons between different CEP methods}

The MTF results in the bottom row of Fig.~\ref{fig:CEP} demonstrate that proposed MoTDiff achieves better invertibility compared to the existing SOTA CEP methods, as indicated by a smaller difference between the maximum and minimum values of the MTF.
The results suggest that code optimization with accurately estimated trajectories yields codes better matched to the ground-truth motions, resulting in better invertibility.

The PSNR and SSIM results in Table~\ref{tab:CEP-deblur-comp} demonstrate that proposed MoTDiff can achieve significantly better deblurring performance compared to existing SOTA CEP methods. 
These results in Table~\ref{tab:CEP-deblur-comp} with those in the middle and bottom rows of Fig.~\ref{fig:CEP} well correspond to the widely known principle that improved invertibility of coded PSF leads to fewer deconvolution artifacts.

\begin{table}[t!]
\centering\small
\caption{
Performance comparisons between different CEP methods (synthetic GoPro test dataset using optimized codes).}
\label{tab:CEP-deblur-comp}

\begin{tabular}{@{}lcccc@{}}
\hline\hline
\multirow{2}{*}{Methods} & \multicolumn{2}{c}{(CEP+deblurring)}
\\
& PSNR $\uparrow$ & SSIM $\uparrow$ 
\\ \hline
W/o CEP     & 19.99 & 0.46 
\\
DNF \cite{dnfcep}  & 24.51 & 0.63 
\\
DCE \cite{dce}     & 24.08 & 0.61 
\\ 
Motion-ETR \cite{ETR} & 24.20 & 0.62 
\\
\textbf{MoTDiff (ours)}       & \textbf{26.19} & \textbf{0.69} 
\\
\hline \hline
\end{tabular}
\end{table}

In comparison with existing SOTA CEP methods that optimize codes using incomplete motion information, 
the proposed MoTDiff framework can successfully optimize codes, thereby achieving good invertibility of coded PSFs.
Existing CEP methods have several limitations, as follows.
DNF \cite{dnfcep} relies solely on the length of a trajectory (i.e., motion speed) assumed to be known, while neglecting directional information of a motion.
DCE \cite{dce} has difficulty optimizing codes for individual motions, 
because it generates only a single code for many input videos. 
Motion-ETR \cite{ETR} faces a challenge in modeling complex motions, as it relies on a parametric representation.

\begin{table}[t!]
\caption{Performance comparisons between different MoTDiff variants (for blind image deblurring, we used the synthetic GoPro test dataset constructed in Section~\ref{sec:data}; for CEP, we used another synthetic GoPro test dataset using optimized codes).}
\label{tab:ablation}
\centering\small

\setlength{\tabcolsep}{3pt}
\begin{threeparttable}

\begin{tabular}{c c c | c c c}
\hline\hline
\multicolumn{6}{c}{(A) Blind Deblurring} \\
\hline
\specialcell{Multi-scale \\ features (\ref{eq:feat-ms})\tnote{a}~} & 
\specialcell{Proposed \\ loss (\ref{eq:overall loss})\tnote{b}~} & 
\specialcell{Proposed \\ STPD (\ref{eq:stpd})} & 
MNC\,$\uparrow$ & 
PSNR\,$\uparrow$ & 
SSIM\,$\uparrow$ \\
\hline
$\times$ ($\mathbf{f}_2$)    & 
$\ocircle$             & $\ocircle$    & 0.28            & 14.59            & 0.44             \\
$\times$ ($\mathbf{f}_4$)    & 
$\ocircle$             & $\ocircle$    & 0.73              & 23.59               & 0.76               \\
$\ocircle$    & 
$\times$             & $\ocircle$    & 0.13            &  8.86            & 0.13             \\
$\ocircle$    & 
$\ocircle$             & $\times$    & 0.76            & 23.73            & 0.77             \\
$\ocircle$    & 
$\ocircle$             & $\ocircle$    & \textbf{0.76}   & \textbf{23.89}   & \textbf{0.77}    \\
\hline\hline
\end{tabular}

\vspace{0.75pc}

\begin{tabular}{c c c | c c}
\hline\hline
\multicolumn{5}{c}{(B) CEP+Deblurring} \\
\hline
\specialcell{Multi-scale \\ features (\ref{eq:feat-ms})\tnote{a}~} & 
\specialcell{Proposed \\ loss (\ref{eq:overall loss})\tnote{b}~} & 
\specialcell{Proposed \\ STPD (\ref{eq:stpd})} & 
PSNR\,$\uparrow$ & 
SSIM\,$\uparrow$ \\
\hline
$\times$ ($\mathbf{f}_2$) & $\ocircle$ & $\ocircle$ & 23.43 & 0.59 \\
$\times$ ($\mathbf{f}_4$) & $\ocircle$ & $\ocircle$ & 24.64 & 0.64 \\
$\ocircle$ & $\times$ & $\ocircle$ & 21.15 & 0.50 \\
$\ocircle$ & $\ocircle$ & $\times$ & 25.02 & 0.65 \\
$\ocircle$ & $\ocircle$ & $\ocircle$ & \textbf{26.19} & \textbf{0.69} \\
\hline\hline
\end{tabular}

\vspace{0.5pc}
\begin{tablenotes}       
\item[a] \parbox[t]{1.15\linewidth}{The first ``$\times$'' setup uses a single-scale feature $\mathbf{f}_2$.
The second ``$\times$'' setup uses a single-scale feature $\mathbf{f}_4$.}
\item[b] \parbox[t]{1.15\linewidth}{The ``$\times$'' setup uses the standard denoising loss in DDPM \cite{ddpm}.}
\end{tablenotes}
\end{threeparttable}
\end{table}

\subsection{Ablation study}

This section evaluates the effectiveness of the three key components of MoTDiff:
\textit{1)} multi-scale feature extraction (\ref{eq:feat-ms}) via the proposed conditioning approach in Section~\ref{subsubsec:feature extraction};
\textit{2)} the proposed training loss (\ref{eq:overall loss}) in Section \ref{sec:loss}; and
\textit{3)} the proposed STPD training strategy (\ref{eq:stpd}) in Section \ref{sec:STPD}.
In Tables~\ref{tab:ablation}(A)--(B), 
the last configuration integrates all the proposed innovations in MoTDiff.

Comparing the results in the first and second rows with those in the fifth row of Table \ref{tab:ablation}(A)--(B) shows that 
the proposed conditioning approach using multi-scale motion features leads to performance improvements in both blind image deblurring and CEP by providing richer conditional information for trajectory estimation.
Among single-scale variants (see the first and second rows in Table \ref{tab:ablation}(A)--(B)), using higher-level features $\mathbf{f}_4$ as guidance for the diffusion model achieves better performance than using intermediate-level features $\mathbf{f}_2$, suggesting that global motion context provides more informative cues for trajectory estimation.

Comparing the results in the third and fifth rows of Table \ref{tab:ablation}(A)--(B) shows that the proposed loss function plays a critical role in improving performance in both blind deblurring and CEP, by promoting fine-grained and spatially consistent trajectory estimation.

Comparing the results of the fourth and fifth rows of Table \ref{tab:ablation}(A)--(B) shows that the proposed STPD strategy is particularly effective for CEP, but less so for for blind image deblurring. 
This is because CEP directly depends on the structural integrity of estimated trajectories, 
where disconnected or fragmented paths can severely hinder code optimization.
In contrast, in blind image deblurring experiments, 
estimated trajectories are resampled into PSFs as described in Section~\ref{sec:debcep}, 
so fragmented trajectories can still yield PSFs similar to those from fully connected trajectories,
provided that the overall trajectory shape is preserved.

\section{Conclusion}
\label{sec:conclusion}

Estimating accurate motion information from a single motion-blurred image is essential in diverse computational imaging and computer vision tasks.
Yet, existing motion representations are often coarse-grained and inaccurate,
which underscores the need for a more expressive and precise motion representation.

In this paper, we proposed HR motion trajectory, 
a new motion representation with significantly higher resolution than existing ones, which can capture complex motion characteristics such as direction and curvature.
We proposed the first diffusion model, MoTDiff, that estimates an HR motion trajectory from a single blurred image. 
The key components of proposed MoTDiff include a novel conditioning approach that provides rich motion information as guidance of diffusion model, and training strategies that effectively ensure spatially coherent and dense trajectory estimation.
Our experimental results demonstrate that using HR trajectories estimated by MoTDiff achieves superior performance improvements on blind deblurring and CEP tasks.

In future work, we plan to design an end-to-end framework that simultaneously estimates HR motion trajectory and performs some end task.

\section*{Acknowledgments}

This work was supported by SKKU Academic Research Support Program, Sungkyunkwan University, 2024.

\bibliographystyle{IEEEtran}
\bibliography{references}

\end{document}